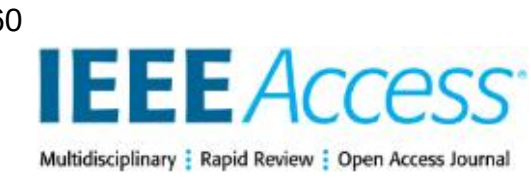

# E-CONAN (Entailment, CONtradition And Neutral) Benchmarks: Arabic Textual Entailment and Natural Inference Datasets.

**Khloud AL Jallad[1], Nada Ghneim[2], and Ghaida Rebdawi[1]**

[1] Higher Institute for Applied Sciences and Technology, Damascus, Syria.
[2] Arab International University, Daraa, Syria.

Corresponding author: Khloud AL Jallad (e-mail: Khloud.aljallad@hiast.edu.sy).

**ABSTRACT** Natural Language Inference (NLI), processes pairs of sentences to extract their semantic relations. NLI has been a hot research topic, integrated as a main component in other NLP applications, from morphological spelling correction tasks to higher-level tasks such as machine translation and information extraction. Despite significant advancements in textual inference across various languages all around the world, Arabic language still suffers from limited resources in this domain, specifically, when it comes to the scarcity of robust well-constructed benchmarks necessary for effective model tuning and generalization. To address this gap, this paper introduces E-CONAN benchmarks that are composed of sentences pairs from various sources: (1) automatically-translated pairs, (2) human-validated machine-translated pairs, (3) hand-crafted pairs from teaching Arabic as foreign language books, and (4) headlines pairs from different news channels containing rumors. E-CONAN contains two benchmark datasets, E-CONAN-2, a 2-way dataset (RTE) and E-CONAN-3, a 3-way dataset (NLI). Additionally, we have used E-CONAN benchmarks to evaluate 9 state-of-the-art multilingual pretrained models using zero-shot classification. Models were evaluated across the ArNLI, XNLI, and E-CONAN datasets. Results show that E-CONAN is a potentially valuable resource for evaluating model generalization and even for fine-tuning pre-trained models. Its diverse composition, derived from a combination of sources, offers a broader and more robust assessment compared to XNLI and ArNLI. Furthermore, results show that mDeBERTa model, pre-trained on 100 languages and fine-tuned on a combination of four machine-translated datasets, demonstrated superior performance on all datasets. It achieved accuracies of 71% and 86% on the E-CONAN-3 and XNLI datasets, respectively, outperforming all other evaluated models. In addition, we have evaluated 5 LLMs on E-CONAN-3 dataset. Best results were achieved by Gemma with an accuracy of 68%. Analyzing these results showed that most errors were between neutral and contradiction classes, thus we calculated results after reformulating E-CONAN-3 using 2-way labeling. Best results are achieved by Gemma and Qwen with an accuracy of 95%, 94% respectively. Moreover, we incorporated MARBERT as a representative Arabic-specific baseline and conducted performance evaluation comparison to demonstrate how Arabic-specific models scale against cross-lingual and LLM-based approaches on the E-CONAN benchmarks. Furthermore, we conducted detailed qualitative and quantitative error analysis to analyze frequent error patterns. Results show that while pretrained models are often misled by lexical overlap, LLMs are often misled by topical familiarity. E-CONAN benchmarks will be publicly available, we hope that it will enrich research community in Arabic textual entailment and natural language inference.

## I. INTRODUCTION

Recognizing Textual Entailment (RTE) is the task of detecting logical relation type between pair of sentences (Text as T and Hypothesis as H). When first introduced, RTE task was called 2-way RTE as it classified pairs into two relations (entail/not entail). The 3-way RTE term appeared in 2008 [1], as the task of determining entailment relation between pairs of sentences introduced three relations (entail/ contradict/ neutral or unknown). Later, the term Natural Language Inference (NLI) has been frequently used for the latter task. The studied semantic relations between the two pair sentences (T and H) are:

1- **Entailment**: Entailment occurs when T and H intersect or when one is contained in the other ($T \cap H \neq \emptyset$ AND $T \cup H$ = Universe) OR (($T \subseteq H$) OR ($H \subseteq T$)). Figure 1(a) represents the Entailment relation using Algebra Sets notation. For instance, the relation between "Sam travelled to Paris" as T and "Sam travelled to France" as H, is T entails H, as T is a part of H. However, the opposite is not right as France or any country have many cities in it.

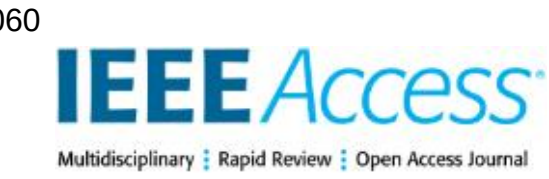

2- **Contradiction**: Contradiction occurs when T and H cannot be true together, i.e., if T is True then H is False and vice versa (T∩H= ∅ AND T∪H=Universe). Figure 1(b) represents the Contradiction using Algebra Sets notation. For instance, the relation between "I like reading" as T and "I hate reading" as H, is contradiction as they cannot be true at the same time.

3- **Neutral**: Neutral occurs when T and H have no semantic relations. Each one of them is a set in the universe, no intersection between them and their union is not the whole universe (T∩H= ∅ AND T∪H ≠ Universe). Figure 1(c) represents the Neutral relation using Algebra Sets notation. The relation between "I like reading" as T and "I hate oranges" as H, is neutral, as there is no contradiction neither entailment between them. The same relation applies if the sentence H was "I like oranges".

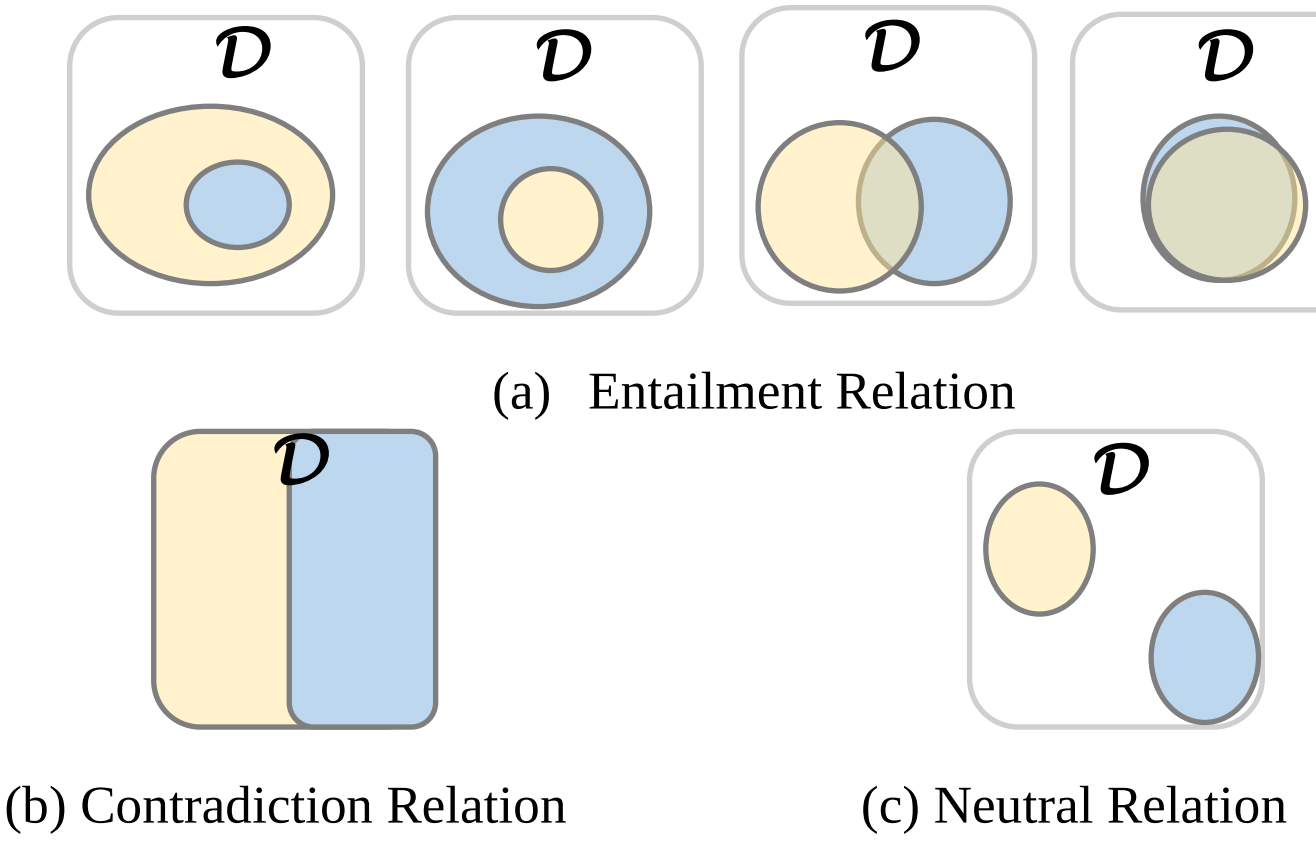

𝓓 is the Universe, Yellow is T, Blue is H.

**FIGURE 1. Pairs Inference Relations Types.**

While significant advancements have been made in textual inference across many languages, Arabic remains under-resourced in this domain, with a lack of high-quality benchmarks that are essential for the proper tuning and generalization of textual inference models. While existing Arabic NLI benchmarks have established a foundational groundwork, they are typically constrained to a single data-generation paradigm. For instance, resources like Arabic XNLI and SNLI rely almost on machine translation, whereas others are strictly limited to a single domain or hand-crafted format. This single constructing paradigm leaves a gap in evaluating models' generalization across real-world linguistic variation. To address this, the proposed E-CONAN benchmarks purposefully unified all structural and linguistic advantages of these varied approaches into a unified benchmark. By merging human-validated machine translations, automatically translated texts, hand-crafted texts, and news headlines, E-CONAN captures a good level of linguistic diversity, spanning formal, casual, translated, and highly nuanced native Arabic structures. Consequently, it provides a uniquely challenging and comprehensive environment to test model generalization across multiple text styles. The main contributions of this paper are:

1- RTE Benchmark (E-CONAN-2 dataset) that contains 24,875 pairs of sentences, collected from ArNLI dataset, the Arabic translated section of SNLI dataset, the Arabic translated section of XNLI dataset, AnsStance Dataset, ArEntail Dataset.

2- NLI Benchmark (E-CONAN-3 dataset) that contains 18,875 pairs of sentences, collected from ArNLI dataset, the Arabic translated section of SNLI dataset, the Arabic translated section of XNLI dataset, AnsStance Dataset.

3- Evaluation of 10 pretrained models to compare E-CONAN datasets with three previous SoTA benchmarks datasets (ArNLI, XNLI, ArEntail).

4- Evaluation of 6 LLMs to compare E-CONAN datasets with three previous SoTA benchmarks datasets (ArNLI, XNLI, ArEntail).

This paper is organized as follows: Introduction is in section 1. Related works are shown in section 2. Section 3 contains the datasets construction process and datasets statistics. Section 4 shows the Evaluation Models. Results and discussions are presented in section 5. Error analysis is shown in section 6, with a conclusion in section 7.

## II. Related Works

RTE was first started as a challenge in PASCAL [2]. PASCAL stands for Pattern Analysis, Statistical Modelling and Computational Learning. PASCAL is a Network of Excellence funded by the European Union. It has established a distributed institute that brings together researchers and students across Europe, and is now reaching out to new countries all over the world. PASCAL run the Recognizing Textual Entailment Challenge series where each of them focus on RTE in specific applications.

Several PASCAL challenges were done by enriching examples each new challenge to represent different levels of entailment reasoning, such as lexical, syntactic, morphological and logical. And most of them focus on four famous NLP applications: Information Retrieval (IR), Information Extraction (IE), Question Answering (QA), and multi-document summarization (SUM) as such systems must recognize the different forms in which their inputs and requested outputs might be expressed, so there is a need to have a framework for logic-based meaning-level representations using RTE. RTE datasets were initially annotated with two classes (entail/not entail), called 2-way RTE. Later, a 3-way RTE emerged with three annotations (entail/contradict/neutral) by splitting the 'not entail' class into 'neutral' if no semantic relation between pairs and 'contradiction' if opposite semantic meaning between them. The 3-way RTE is also called Natural Language Inference (NLI).

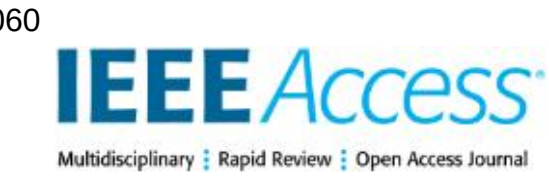

Table 1 shows comparisons between datasets of PASCAL RTE challenges.

TABLE I
PASCAL RTE CHALLENGE RTE DATASETS

| | Year | Description | Main Task | Main Sub tasks | Annotations |
|---|---|---|---|---|---|
| **RTE 1 [2]** | 2006 | It contains manually collected pairs of Text-Hypothesis. Hypothesis (H) length is same as Text(T) length and it was 1-2 sentences. | IE, IR, QA, SUM | - | 2-way |
| **RTE 2 [3]** | 2006 | It was divided into training and testing set. Then it was divided into 200 text-hypothesis pairs for different applications. | IE, IR, QA, SUM | - | 2-way |
| **RTE 3 [4]** | 2007 | No major changes on the dataset | IE, IR, QA, SUM | - | 2-way |
| **RTE 4 [5]** | 2008 | It was the first conference paper that proposed RTE as three-judgment task where they added the class contradiction to the labels to express if two sentences do not entail same meaning but also entail contradicted meanings. The “Text” was longer than the text in RTE-3 where the “Hypothesis” length was the same. | IR, QA, SUM | - | 3-way |
| **RTE 5 [6]** | 2009 | The added value was that Textual Entailment judgment is done over a real corpus of Summarization scenario | Search Pilot | - | 3-way |
| **RTE 6 [7]** | 2009 | Saves the same features of RTE-5 | Update Summarization on scenario | Novelty Detection And KBP Validation Pilot | 3-way |
| **RTE 7 [8]** | 2011 | Saves the same features of RTE-5 and RTE-6 | Update Summarization on scenario | Novelty Detection and KBP Slot Filling | 3-way |

In English, numerous NLI datasets were published in the last few years, such as WNLI [9], SNLI [10], MultiNLI [11]. Additionally, GLUE proposed RTE dataset [12] which is a combination of RTE1 [2], RTE2 [3], RTE3 [4], and RTE5 [6]. Some of these datasets were automatically translated to 15 languages including Arabic, such as XNLI [13] and SNLI [14]. However, this automatic translation introduced errors in meaning of pairs that affects labels and thus models' performance.

Using these datasets, many state-of-the-art models were proposed such as: PaLM 540B[15], Vega_v2_6B[16], RoBERTa[17], SemBERT[18], XLNET[19], AlexaTM_20B[20], BloombergGPT and GPT-NeoX and BLOOM_176B [21], UnitedSynT5[22], ByT5[23] , Rethinking Coupling[24], mGPT[25], DeBERTa[26], SpanBERT[27], SqueezeBERT[28], DistilBERT[29].

As for Arabic language, there are still few datasets, as follows:

- 2-Way ArbTEDS [30] that consists of 618 text-hypothesis pairs collected from Arabic news websites (Al Jazeera, Al Arabiya, BBC Arabic)[1] and from annotated pairs collected by hand. These pairs cover a number of domains such as politics, business, sport and general news. ArbTEDS was annotated by eight expert/non-expert volunteer annotators.
- 2-Way ArEntil Dataset [31] that contains 6000 sentence pairs collected from news headlines and manually labeled.
- 3-Way ArNLI [32] that contains 6366 pairs divided as (1932 entailment, 1073 contradiction, and 3361 neutral).

Some state-of-the-art models that contain Arabic language are: Facebook Bart Large MNLI [33], [34], Facebook RoBERTa-Large-MNLI [35] , Microsoft Multilingual MiniLM [36], Microsoft XLM-RoBERTa

[1] Al Jazeera http://www.aljazeera.net/ Al Arabiya http://www.alarabiya.net/ and BBC Arabic http://www.bbc.co.uk/arabic/

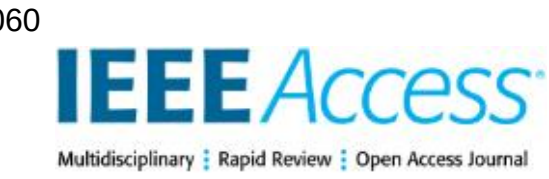

[36], Microsoft DeBERTaV3 [37]. In addition, Laurer et al. tuned many versions of pretrained multilingual models such as, mDeBERTa-V3-Base-XNLI-Multilingual-NLI-2mil7 [38],Moritzlaurer Ernie-M-Base-MNLI-XNLI [38], Moritzlaurer Multilingual-MiniLMv2-L6-MNLI-XNL [38], Moritzlaurer Multilingual-MiniLMv2-L12-MNLI-XNLI[38]. We have used many of them as zero-shot classification to test and validate our proposed datasets.

Recently, Natural Language Inference and cross-lingual inference have become critical tasks for benchmarking the reasoning capabilities of Large Language Models (LLMs). Recent studies have shifted from pre-trained models to evaluating LLMs, such as Allam [46] , Qwen [47], [48], Command R7B Arabic [53], DeepSeek-R1 [54] and Gemma [52] on complex semantic reasoning. However, evaluating Arabic NLI still faces a significant challenge due to the scarcity of high-quality, native Arabic NLI datasets, with many existing benchmarks relying on machine translation. Our work directly addresses this gap by introducing two Arabic NLI datasets and establishing a comprehensive evaluation framework across 5 cutting-edge LLMs and 10 foundational pre-trained models.

## III. E-CONAN Datasets

### a. *DATASETS CONSTRUCTION*

We have created two datasets: a 2-way dataset (RTE) named E-CONAN-2 and a 3-way dataset (NLI) named E-CONAN-3. These datasets were collected from several sources as follows:

1- ArNLI dataset [32] which contains:
- Manually collected pairs.
- ArbTEDS dataset[2] [30] which contains news collected from Arabic news channels, labeled manually by eight annotators, where an annotator agrees with at least one co-annotator (average around 91% between annotators.
- Automatically translated pairs from English NLI datasets (SICK3 [39], PHEME4 [40], and Stanford Real Life contradiction corpus[5][41]), then manually verified. Where the SICK inter–rater agreement was 84%, as an average, 84% of participants agreed with the majority vote in each pair.

2- The Arabic-translated section of SNLI dataset[6] [14], which contains the first 1,332 test pairs of SNLI dataset [10], manually translated by human experts. Each pair was validated by five independent annotators. It achieved a consensus rate of 98% for three annotators and a 58% consensus rate from all five annotators.

3- The Arabic-translated section of XNLI dataset [42], which was constructed as an evaluation set for Cross-lingual Language Understanding (XLU) by extending the development and test sets of the Multi-Genre Natural Language Inference Corpus (MultiNLI) to 15 languages, including low-resource languages such as Arabic. To ensure transparency regarding data quality, the authors reported specific translation metrics: an Ar-En BLEU score of 35.2, an En-Ar BLEU of 15.8, and a Word Translation P@1 of 51.9. Each pair was validated by five independent annotators. It achieved a consensus rate of 93% for three annotators.

4- AnsStance Dataset [43]: A previous work, which involved transforming stance datasets into Natural Language Inference (NLI) datasets, indicated that AnsStance's language complexity and word counts align closely with our collection. For AnsStance dataset inclusion, we considered s1, s2 as Text and Hypothesis, respectively.

5- ArEntail Dataset [31], contains sentence pairs collected from Arabic news headlines and manually labelled to indicate entailment / not entailment. This source was only used in the E-CONAN-2 dataset.

### b. *Datasets Annotation Mapping*

As for E-CONAN-3, the original NLI annotations were preserved without modification. To ensure consistency across combined datasets, we unified the AnsStance pair annotations by mapping them to the standard NLI labels, where the pairs were reannotated as follows:
- "agree" class was converted to "entailment" class.
- "disagree" class was converted to "contradiction" class.
- all remaining classes were converted to "neutral" class.

[2] "Arabic Textual Entailment Dataset", 2013: http://www.cs.man.ac.uk/~ramsay/ArabicTE/
[3] "SemEval2014", 2014: http://alt.qcri.org/semeval2014/
[4] "Pheme," 2016: https://www.pheme.eu/2016/04/12/pheme-rte-dataset/
[5] "Stanford Real Life contradiction corpus", 2008: https://nlp.stanford.edu/projects/contradiction/
[6] "Arabic SNLI Dataset", 2018: https://bitbucket.org/nlpitu/xnli/src/master/

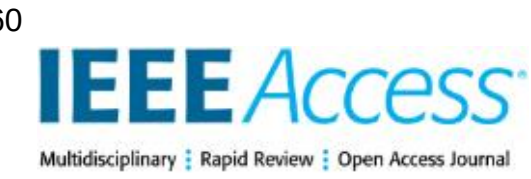

As for E-CONAN-2, All 2-way sources were preserved without modification. For sources containing 3-way annotations, we performed a label transformation to align them with a binary 2-way classification. The mapping was conducted as follows:

- "entail" class was preserved without modification.
- "contradiction" class and "neutral" class were merged in “not-entail” class.

A workflow diagram is shown in Figure 2 that illustrates the full pipeline from raw data sources to the final unified datasets.

Although source datasets were independently validated in their original studies, we acknowledge that inter-annotator agreement was not re-evaluated post-merging. This absence remains a limitation of this study.

### *C. Datasets Statistics*

E-CONAN-2 dataset contains 37% of samples in the class entail, 63% in the class not-entail. While, E-CONAN-3 dataset contains 32% of samples in the class entailment, 34% in the class neutral, and 34% in the class contradiction. Number of samples for each class in each dataset are shown in Table 2.

Statistics on lexical characteristics in Text and Hypothesis are shown in Table 3. It shows a minimum of 2 words by sentence and a maximum of 39 and 59 words in Text, Hypothesis, respectively. The average length of sentence is approximately 8 and 11 in Text, Hypothesis, respectively. This broad range of sentence lengths indicates the presence of both short and concise expressions as well as longer, information-rich constructions, contributing to lexical and structural diversity. The similarity of these statistics across E-CONAN-2 and E-CONAN-3 also demonstrates consistency in dataset construction while preserving linguistic variability.

Figure 3 illustrates the number of samples from each of our data sources. Notably, the highest number of samples are from XNLI and ArNLI. The lowest number of samples is from SNLI. ArEntail is exclusively used in the E-CONAN-2 dataset due to its binary classification of entailment versus non-entailment, as it does not categorize non-entailment into contradiction or neutral.

Figure 4 shows the domain distribution & semantic coverage of E-CONAN datasets, where the datasets span four distinct text styles: News (35%), Simple Everyday (30%), Social Media (20%), and Multi-Genre (15%), compiled from eight source datasets, including ArbTEDS, ArEntail, SNLI, SICK, PEHEM, AnsStance, XNLI, and Stanford Contradiction. The News domain (35%) contributes formal vocabulary, relatively complex syntactic structures, and information-dense content. The Simple Everyday Text domain (30%) provides straightforward lexical patterns suitable for evaluating baseline reasoning capabilities. Social Media data (20%) introduces informal language, while the Multi-Genre domain (15%) expands cross-domain semantic coverage. Collectively, this heterogeneous composition promotes lexical diversity and broad semantic representation, reducing the likelihood of models adapting to a narrow writing style or vocabulary.

Figure 5 shows the word cloud of E-CONAN-2 and E-CONAN-3, highlighting that it contains words from diverse domains such as geographical words (“السعودية”,”الولايات المتحدة”, “إيران”, “سوريا”…), political words ( “المعارضة”, “داعش”, …), economical words (“ذهب”, “دولار”, “أسعار”, …), general words (“رجل”, “امرأة”, “أطفال”, “لعب”, “يرتدي”, “حديقة”, …).

TABLE 2
E-CONAN DATASETS CLASSES DISTRIBUTION

| Dataset | Class | # Samples |
|---|---|---|
| E-CONAN-2 | Entail | 9157 |
| | Not-Entail | 15718 |
| | Total Pairs | 24875 |
| E-CONAN-3 | Entailment | 6157 |
| | Contradiction | 6357 |
| | Neutral | 6361 |
| | Total Pairs | 18875 |

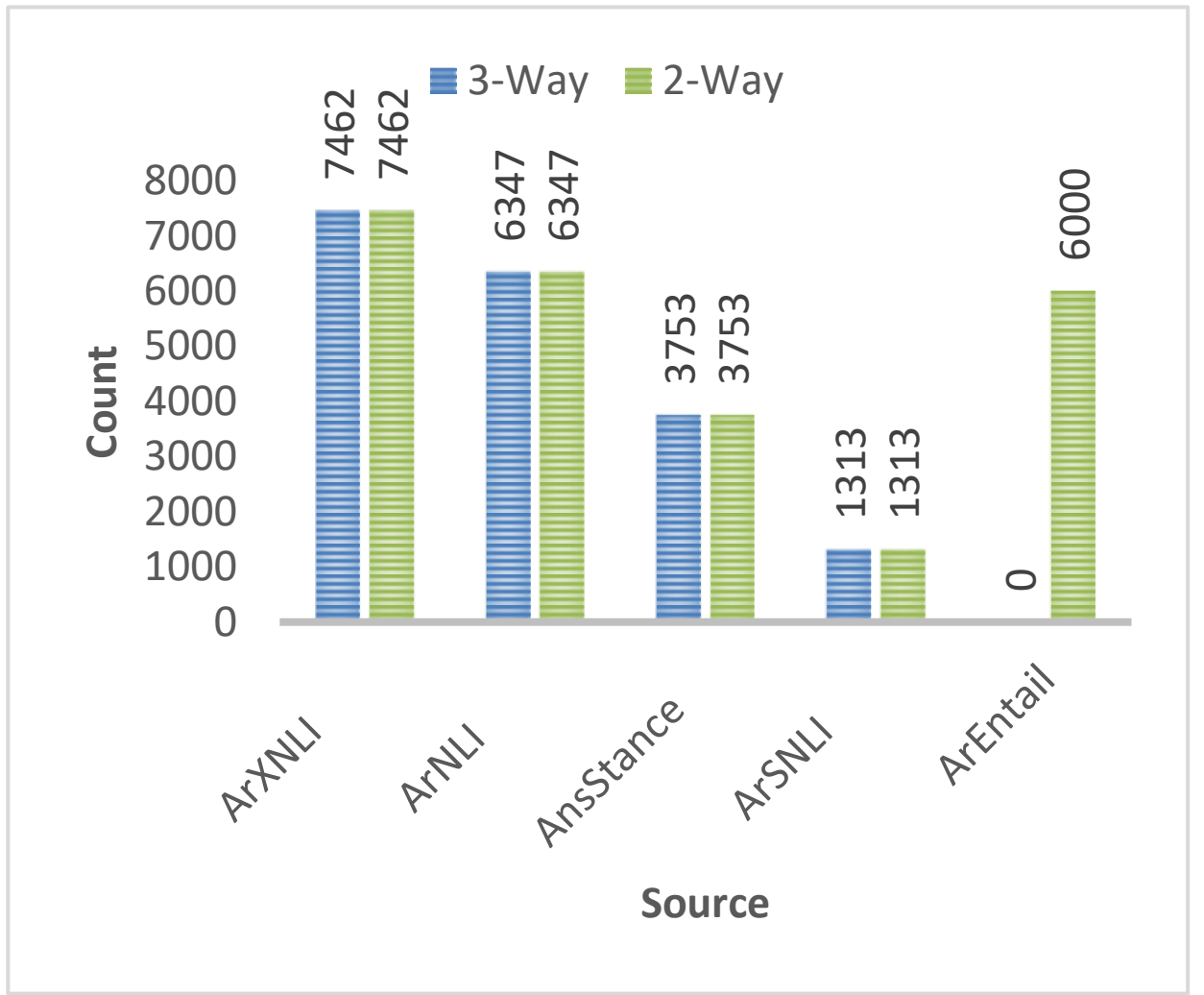

**Figure3: E-CONAN Datasets Statistics of Sources Distribution**

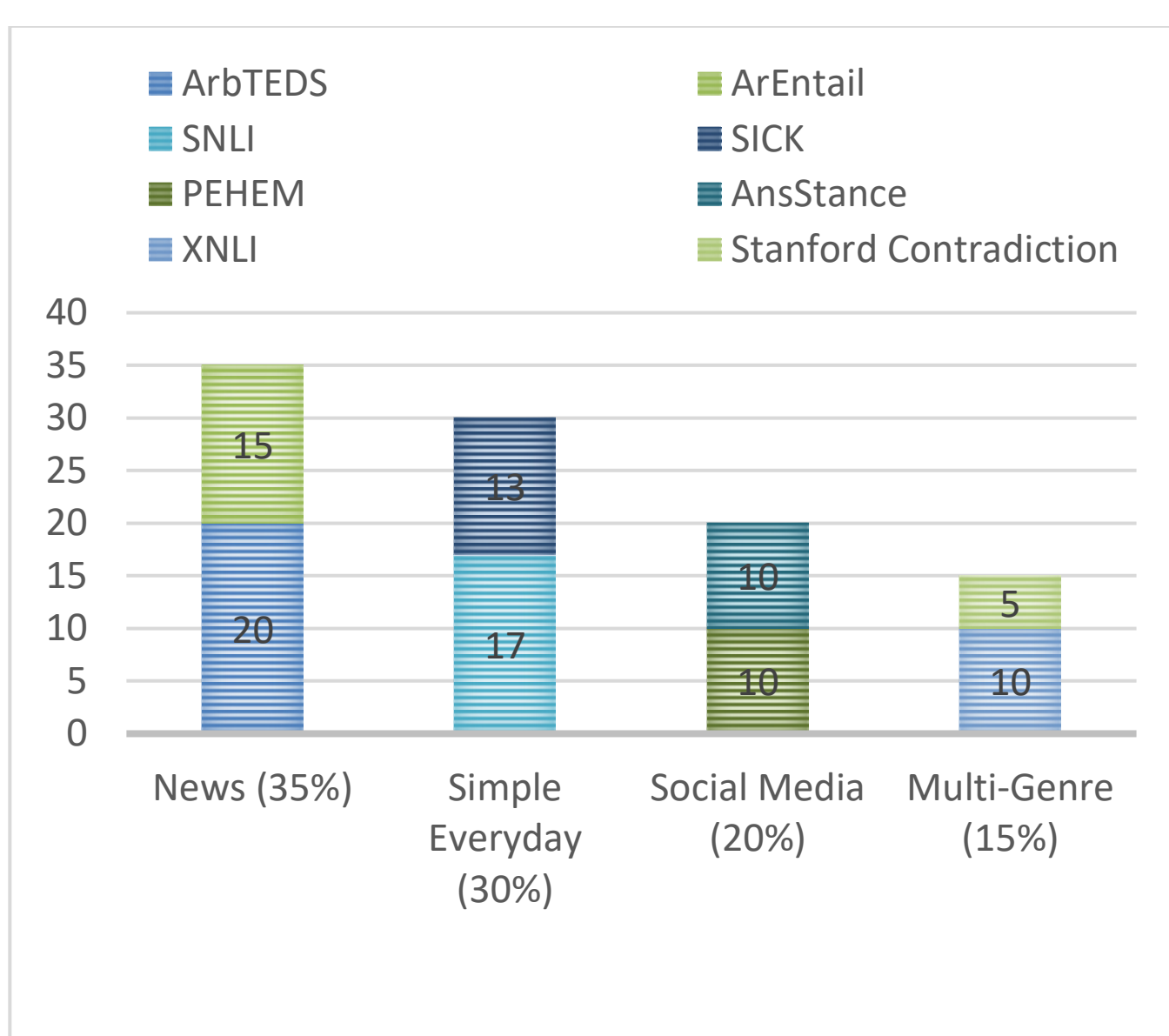

***Figure4:*** **E-CONAN Datasets Statistics of Domain Distribution**

We have conducted quantitative comparisons of state-of-the-art (SoTA) benchmarks with E-CONAN-2, E-CONAN-3 benchmarks, as presented in Table 4 and Table 5, respectively. As shown in Table 4 and Table 5, our datasets maintain a robust distribution across classes. Notably, E-CONAN-3 achieves a highly optimized, near-perfect balance across the three standard NLI classes (Entailment: ~32.6%, Contradiction: ~33.7%, Neutral: ~33.7%), resolving the distribution skews found in older benchmarks like ArNLI.

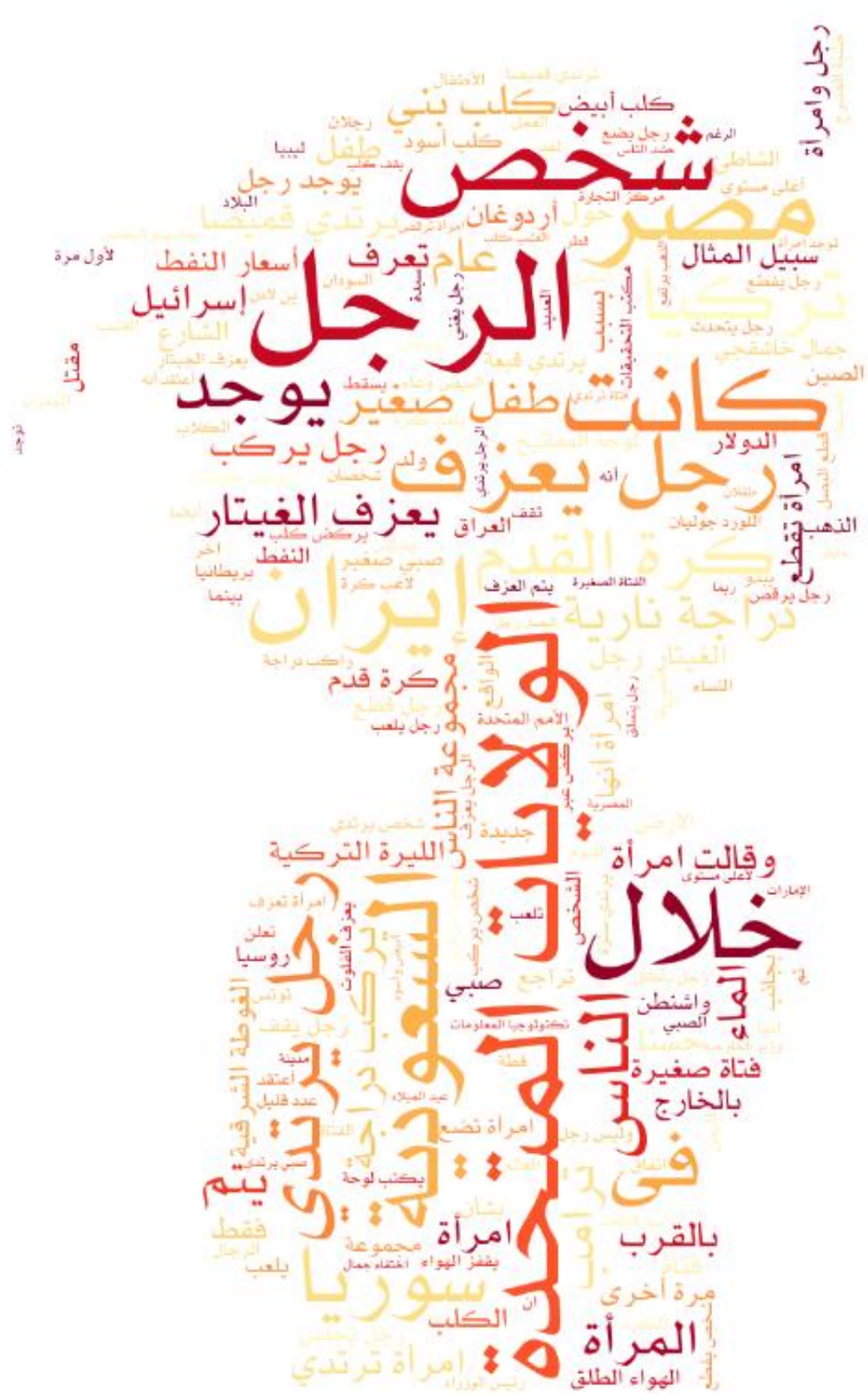

**Figure5: E-CONAN Word Cloud**

TABLE 3
LEXICAL CHARACTERISTICS IN E-CONAN-2 AND E-CONAN-3

| | E-CONAN-2 | | E-CONAN-3 | |
|---|---|---|---|---|
| | **T** | **H** | **T** | **H** |
| **Max** | 39 | 59 | 39 | 59 |
| **Average** | 8.57 | 11.62 | 8.04 | 11.37 |
| **Min** | 2 | 2 | 2 | 2 |

TABLE 4
QUANTITATIVE COMPARISONS OF E-CONAN-2 WITH SOTA BENCHMARK

| Class | ArEntail | E-CONAN-2 |
|---|---|---|
| **Entail** | 3000 | 9157 |
| **Not-Entail** | 3000 | 15718 |
| **Total pairs** | 6000 | 24875 |

TABLE 5
QUANTITATIVE COMPARISONS OF E-CONAN-3 WITH SOTA BENCHMARK

| Class | ArNLI | ArSNLI | ArXNLI Test + Validation | E-CONAN-3 |
|---|---|---|---|---|
| **Entailment** | 1928 | 450 | 2485 | 6157 |
| **Contradiction** | 1068 | 429 | 2486 | 6357 |
| **Neutral** | 3351 | 434 | 2491 | 6361 |
| **Total pairs** | 6347 | 1313 | 7462 | 18875 |

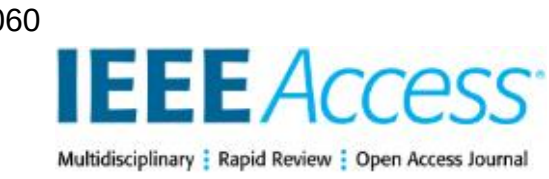

Beyond dataset size, the statistics of E-CONAN datasets indicate substantial linguistic diversity, semantic coverage, and class balance. The source distribution analysis reveals that E-CONAN integrates data from eight datasets spanning four domains: News, Simple Everyday Text, Social Media, and Multi-Genre. This heterogeneous composition promotes lexical diversity and broad semantic representation. In addition, E-CONAN incorporates sentence pairs originating from multiple construction paradigms, including automatically translated pairs, human-validated machine-translated pairs, hand-crafted educational examples, and news headline pairs involving rumors and factual inconsistencies. These diverse data-generation processes further broaden the linguistic and semantic coverage of the benchmark.
The lexical statistics indicate substantial variation in sentence lengths, ranging from short expressions to longer information-rich constructions. This variability contributes to lexical and structural diversity while maintaining consistency across E-CONAN-2 and E-CONAN-3.
Finally, the class distributions demonstrate improvements over several existing Arabic NLI benchmarks. In particular, E-CONAN-3 maintains a highly balanced distribution across the Entailment, Contradiction, and Neutral classes, providing a more balanced benchmark for evaluating semantic alignment and inference performance.

## IV. Evaluation Models

### a. Pretrained Models

We used our dataset E-CONAN to evaluate some famous state-of-the-art pretrained models that cover Arabic language without any training examples (zero-shot classification), as shown in Figure 6. Most of these pretrained models were trained on MNLI only or on both MNLI and XNLI.
We used the following models:

- Facebook Bart Large MNLI[7] [33], [34].
- Moritzlaurer MiniLM-L6-MNLI8: Microsoft Multilingual MiniLM [36] tuned on MNLI dataset [11].
- Moritzlaurer Deberta-V3-Base-MNLI[9]: Microsoft DeBERTaV3 [37] tuned on MNLI dataset [11].
- Moritzlaurer-mDeBERTa-V3-Base-MNLI-XNLI[10][38]: a tuned version of a Microsoft mDeBERTa-v3-base [37] model, which was trained on CC100 multilingual dataset [44][45] with 100 different languages, tuned on both MNLI [11] and XNLI [42] datasets.
- Moritzlaurer-mDeBERTa-V3-Base-XNLI-Multilingual-NLI-2mil7[11] [38]: a tuned version of a Microsoft mDeBERTa-v3-base [37] model, which was trained on CC100 multilingual dataset [44][45], tuned on both MNLI [11] and multilingual-NLI-26lang-2mil7 [38]datasets.
- FacebookAI RoBERTa-Large-MNLI[12][35]: which was trained on MNLI dataset [11].
- Moritzlaurer Ernie-M-Base-MNLI-XNLI[13] [38]: a tuned version of Meta's RoBERTa model, tuned on both MNLI [11] and XNLI [42] datasets.
- Moritzlaurer Multilingual-MiniLMv2-L6-MNLI-XNLI[14] [38]: a Microsoft XLM-RoBERTa [36] model tuned on both MNLI [11] and XNLI [42] datasets.
- MoritzlaurerMultilingual-MiniLMv2-L12-MNLI-XNLI[15][38]: a Microsoft XLM-RoBERTa [36] model tuned on both MNLI [11] and XNLI [42] datasets.

Furthermore, to evaluate the performance of cross-lingual models against Arabic pre-trained models, we utilized a version[16] of MARBERT [46] that was fine-tuned on the Arabic XNLI dataset (as illustrated in Figure 7). Specifically, we conducted experiments to compare the results of Arabic-specific pre-trained models with those of multilingual models.

### b. LLMs

Furthermore, we used our dataset E-CONAN to evaluate some famous state-of-the-art large language models that cover Arabic language, using zero-shot classification (See Figure 8). The comparison was based on the following LLMs:

1- **ALLAM** [46] ALLaM stands for Arabic Large Language Model, a series of large language models to support the ecosystem of Arabic Language Technologies (ALT). ALLAM models are based on an autoregressive decoder-only architecture and are pretrained on a mixture of Arabic and English texts via vocabulary expansion.
2- **Qwen2.5** [47], [48] is a comprehensive series of large language models (LLMs) developed by Alibaba Cloud, based on a Transformer-based decoder-only architecture. The series includes both open-weight dense models (0.5B to 72B parameters) and proprietary Mixture-of-Experts

[7] https://huggingface.co/facebook/bart-large-mnli
[8] https://huggingface.co/MoritzLaurer/MiniLM-L6-mnli
[9] https://huggingface.co/MoritzLaurer/DeBERTa-v3-base-mnli
[10] https://huggingface.co/MoritzLaurer/mDeBERTa-v3-base-mnli-xnli
[11] https://huggingface.co/MoritzLaurer/mDeBERTa-v3-base-xnli-multilingual-nli-2mil7
[12] https://huggingface.co/FacebookAI/roberta-large-mnli
[13] https://huggingface.co/MoritzLaurer/ernie-m-base-mnli-xnli
[14] https://huggingface.co/MoritzLaurer/multilingual-MiniLMv2-L6-mnli-xnli
[15] https://huggingface.co/MoritzLaurer/multilingual-MiniLMv2-L12-mnli-xnli
[16] https://huggingface.co/vish88/MARBERTv2-xnli-finetuned

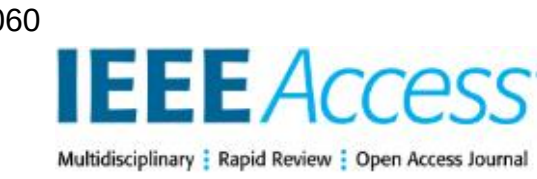

(MoE) variants (Qwen2.5-Turbo and Qwen2.5-Plus).

3- **Gemma3** [52]: an open-weight, multimodal model developed by Google. Ranging in size from 1 to 27 billion parameters. It has 128K-token context window. It supports multilingual (140+ languages) and multimodal input (text and images).

4- **Command R7B Arabic** [53] 7-billion parameter, open-weights, multilingual LLM by Cohere Labs The model has been trained and evaluated for performance in Arabic and English, but its training data includes samples from other languages. Command R7B Arabic supports a context length of 128,000 tokens. It has been trained specifically for tasks such as the generation step of Retrieval Augmented Generation (RAG) in Arabic and English.

5- **DeepSeek-R1** [54]: A large-scale, open-weight Mixture-of-Experts model from DeepSeek AI (with 671B total parameters and ≈37B active per query). It is optimized for complex reasoning and logical tasks (e.g., math, coding, scientific reasoning). The maximum generation length is set to 32,768 tokens. Its pipeline incorporates two RL stages aimed at discovering improved reasoning patterns and aligning with human preferences.

### c. *Experiments Setup:*

We conducted all experiments using a zero-shot inference. To ensure fairness, transparency, and reproducibility, all models were evaluated using the same configuration without fine-tuning, few-shot examples, or task-specific adaptations. The decoding parameters were maintained strictly at their standard pre-configured default settings across all models. Specifically, we used a temperature of 0, and disabled penalty restrictions to guarantee deterministic and consistent outputs.

As for LLMs prompt, the inference was driven by a structured zero-shot prompt command provided to the models together with each text pair. The exact prompt template employed for the evaluation is formulated as follows: "Having the following sentences Text: {t} and Hypothesis {h} Answer: Let's classify the relation as one of the following classes ['entailment', 'contradiction', 'neutral']."

## V. Results and Discussion

As agreed in General Language Understanding Evaluation benchmark (GLUE) [12], the evaluation metrics for NLI is accuracy. For this reason, accuracy is the main metric that is used in all experiments.

### a. *Pretrained Results*

Accuracy percentages of cross-lingual pretrained models zero-shot classification results are shown for all studied datasets in Tables 6, 7. Macro precision, recall and F1 results are illustrated in Figures 9, 10 on ArNLI, E-CONAN and XNLI datasets respectively.

TABLE 6
ACCURACY COMPARISONS WITH SOTA NLI ZERO-SHOT CLASSIFICATION

| Model Name | ArXNLI Test | ArXNLI Test + Validation | ArNLI | E-CONAN-3 |
|---|---|---|---|---|
| FacebookAI/roberta-large-mnli | 0.37 | 0.36 | 0.23 | 0.34 |
| Facebook/bart-large-mnli | 0.38 | 0.38 | 0.25 | 0.36 |
| MoritzLaurer/mDeBERTa-v3-base-xnli-multilingual-nli-2mil7 | 0.79 | **0.86** | **0.55** | **0.71** |
| MoritzLaurer/MiniLM-L6-mnli | 0.35 | 0.35 | 0.26 | 0.33 |
| MoritzLaurer/mDeBERTa-v3-base-mnli-xnli | **0.80** | 0.87 | 0.49 | 0.70 |
| MoritzLaurer/multilingual-MiniLMv2-L6-mnli-xnli | 0.69 | 0.77 | 0.46 | 0.61 |
| MoritzLaurer/DeBERTa-v3-base-mnli | 0.57 | 0.57 | 0.37 | 0.46 |
| MoritzLaurer/multilingual-MiniLMv2-L12-mnli-xnli | 0.73 | 0.81 | 0.44 | 0.64 |
| MoritzLaurer/ernie-m-base-mnli-xnli | 0.78 | 0.85 | 0.48 | 0.68 |

TABLE 7
ACCURACY COMPARISONS WITH SOTA RTE ZERO-SHOT CLASSIFICATION

| Model Name | ArEntail | E-CONAN-2 |
|---|---|---|
| FacebookAI/roberta-large-mnli | 0.55 | 0.5 |
| Facebook/bart-large-mnli | 0.56 | 0.66 |
| MoritzLaurer/mDeBERTa-v3-base-xnli-multilingual-nli-2mil7 | 0.69 | 0.84 |
| MoritzLaurer/MiniLM-L6-mnli | 0.53 | 0.47 |
| MoritzLaurer/mDeBERTa-v3-base-mnli-xnli | **0.72** | **0.83** |
| MoritzLaurer/multilingual-MiniLMv2-L6-mnli-xnli | 0.64 | 0.74 |
| MoritzLaurer/DeBERTa-v3-base-mnli | 0.55 | 0.6 |
| MoritzLaurer/multilingual-MiniLMv2-L12-mnli-xnli | 0.67 | 0.77 |
| MoritzLaurer/ernie-m-base-mnli-xnli | 0.67 | 0.81 |

As for datasets comparisons, we note that lowest results were on ArNLI dataset. ArNLI is probably the hardest dataset as it only contains complex sentences that were carefully checked by humans. These sentences were either manually translated or manually created from scratch. On the other hand, the highest results were on XNLI dataset. This is likely due to the fact that many models were trained or fine-tuned on XNLI, or on similar machine-translated pairs. The E-CONAN datasets results were somewhere in the middle, as they include a mix of sentences: a part that contains machine-translated pairs checked by humans, an automatically translated part, a part collected from news articles and a part created from scratch by humans. As E-CONAN datasets offer this variety, we believe that it could be a good base to test how well models can generalize, or even to improve pre-trained models.

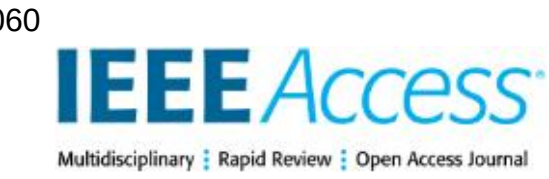

As for models' evaluation comparisons, results show that the mDeBERTa model, pre-trained on 100 languages and fine-tuned on a combination of four machine-translated datasets, outperformed all other models on all datasets. Additionally, results show that all multilingual pre-trained models tuned both on XNLI and MNLI perform better than models tuned on MNLI alone.

### *b. LLM Results:*

Results comparisons of the used LLMs results are shown in Table 8. Where most best results were achieved by Gemma3 on most datasets. Best result on AnsStance dataset was achieved by Gemma3 with an accuracy of 86% and lowest result was by Allam with an accuracy of 68%.

Best result on ArNLI dataset was achieved by Qwen2.5 with an accuracy of 64% and lowest result was by Allam with an accuracy of 50%. Best result on AnsStance dataset was achieved by Gemma3 with an accuracy of 86% and lowest result was by Allam with an accuracy of 68%.

Best result on XNLI dataset was achieved by Gemma3 with an accuracy of 68% and lowest result was by Allam with an accuracy of 56%. Best result on SNLI dataset was achieved by Gemma3 with an accuracy of 48% and lowest result was by Allam with an accuracy of 32%. Best result on E-CONAN-3 dataset was achieved by Gemma3 with an accuracy of 68% and lowest result was by Allam with an accuracy of 55%.

As for results on E-CONAN-3 as 2-way, best result was achieved by Gemma3, Qwen2.5 with an accuracy of 95%, 94% respectively, and lowest result was by Allam with an accuracy of 66%.

TABLE 8
COMPARISON OF LLM ZERO-SHOT CLASSIFICATION ACCURACY ON E-CONAN-3 DATASET & ALL ITS SOURCES

| LLM | AnsStance | ArNLI | ArXNLI Test + Validation | SNLI | E-CONAN-3 | E-CONAN-3 as 2-way |
|---|---|---|---|---|---|---|
| **Allam** | 0.68 | 0.5 | 0.56 | 0.32 | 0.55 | 0.66 |
| **Command R7B Arabic** | 0.79 | 0.64 | 0.61 | 0.4 | 0.64 | 0.92 |
| **DeepSeek** | 0.67 | 0.6 | 0.62 | 0.41 | 0.61 | 0.87 |
| **Gemma3** | 0.86 | 0.61 | 0.68 | 0.48 | 0.68 | **0.95** |
| **Qwen2.5** | 0.71 | 0.64 | 0.64 | 0.38 | 0.64 | **0.94** |

Macro precision, recall and F1 results of all used LLM on our created benchmarks and SoTA benchmarks are illustrated in Figures 11, 12, and 13.

When analyzing results, we notice that most errors were cause by the confusion between neutral and contradiction classes. Consequently, we did an experiment converting 3-way dataset to 2-way dataset (by merging contradiction and neutral into 'not-entail') and the obtained accuracy increased by at least 15% to 30%. Where Allam accuracy increased from %55 to %66, Command R7B Arabic accuracy increased from %64 to %92, DeepSeek accuracy increased from %61 to %87, Gemma3 accuracy increased from %.68 to %95, and Qwen2.5 accuracy increased from %64 to %94.

Moreover, we noticed that highest results were achieved on AnsStance and the lowest results were on SNLI. We think that this might be related to the pretrained dataset domain, as most of our studied models were pretrained on much more news data compared to logical data. In summary, the results highlight the importance of diverse, high-quality benchmarks for accurately assessing and improving model generalization, particularly in resource-limited languages such as Arabic. We do hope that future researches make use of E-CONAN datasets to further refine multilingual models to enhance RTE and NLI performance in diverse linguistic contexts.

Additionally, we incorporated MARBERT as a representative Arabic-specific baseline to enrich our comparative analysis. This shows a performance evaluation comparison between an Arabic-centric model, the top-performing cross-lingual model (mDeBERTa-v3-base-xnli multilingual-nli-2mil7), and the top-performing LLM(Gemma). As detailed in Tables 9 and 10, this comparison demonstrates how Arabic-specific models scale against cross-lingual and LLM-based approaches on the E-CONAN benchmarks. Our results indicate that Gemma achieved the highest performance across the majority of the datasets. Additionally, we observed that performance on 2-way classification significantly outperformed 3-way results; most errors in the 3-way setting occurred between the "neutral" and "contradiction" classes, which are merged into "not-entailment" in the 2-way configuration.

Regarding specific datasets, Gemma provided the best results on the most challenging dataset (ArNLI), outperforming both Arabic-specific and cross-lingual pretrained models by a significant margin. Conversely, mDeBERTa-v3-base-xnli-multilingual-nli-2mil7 yielded the best results on the ArXNLI dataset. This superior performance may be due to the model's prior training on MNLI in addition to XNLI, which likely contributed to its high scores on the E-CONAN-3 dataset.

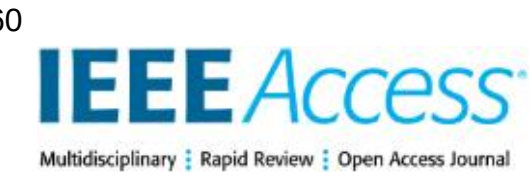

### c. McNemar's Test

To confirm the validity of our zero-shot evaluation shown in Tables 9, 10 on all used datasets, we performed a pairwise McNemar's test with a Bonferroni correction for multiple comparisons. The statistical test comparing Gemma3 against mDeBERTa-v3-base-xnli-multilingual-nli-2mil7on all datasets yielded a p-value <0.000001. This extremely low p-value confirms that the performance differences are highly statistically significant and mathematically meaningful.

TABLE 9
COMPARATIVE ANALYSIS OF REPRESENTATIVE MODELS ON E-CONAN-3 AND SOTA BENCHMARKS

| Dataset Name | Model | Accuracy | Precision | Recall | F1 |
|---|---|---|---|---|---|
| **ArNLI** | **MarBERT-XNLI-Tuned** | 0.44 | 0.49 | 0.55 | 0.44 |
| | **mDeBERTa-v3-base-xnli-multilingual-nli-2mil7** | 0.55 | 0.55 | 0.59 | 0.54 |
| | **Gemma3** | **0.68** | **0.68** | **0.69** | **0.67** |
| **ArXNLI Test + Validation** | **MarBERT-XNLI-Tuned** | 0.71 | 0.73 | 0.71 | 0.72 |
| | **mDeBERTa-v3-base-xnli-multilingual-nli-2mil7** | **0.86** | **0.87** | **0.86** | **0.86** |
| | **Gemma3** | 0.68 | 0.68 | 0.69 | 0.67 |
| **ArSNLI** | **MarBERT-XNLI-Tuned** | 0.41 | 0.42 | 0.41 | 0.41 |
| | **mDeBERTa-v3-base-xnli-multilingual-nli-2mil7** | 0.40 | 0.47 | 0.40 | 0.34 |
| | **Gemma3** | **0.48** | **0.48** | **0.58** | **0.49** |
| **AnsStance** | **MarBERT-XNLI-Tuned** | 0.79 | 0.62 | 0.77 | 0.63 |
| | **mDeBERTa-v3-base-xnli-multilingual-nli-2mil7** | 0.77 | 0.64 | 0.74 | 0.61 |
| | **Gemma3** | **0.86** | **0.91** | **0.69** | **0.77** |
| **E-CONAN-3** | **MarBERT-XNLI-Tuned** | 0.62 | 0.61 | 0.62 | 0.61 |
| | **mDeBERTa-v3-base-xnli-multilingual-nli-2mil7** | **0.71** | **0.72** | **0.71** | **0.71** |
| | **Gemma3** | 0.68 | 0.68 | 0.67 | 0.67 |

TABLE 10
COMPARATIVE ANALYSIS OF REPRESENTATIVE MODELS ON E-CONAN-2 AND SOTA BENCHMARKS

| Dataset Name | Model | Accuracy | Precision | Recall | F1 |
|---|---|---|---|---|---|
| **ArEntail** | MarBERT-XNLI-Tuned | 0.70 | 0.74 | 0.70 | 0.68 |
| | mDeBERTa-v3-base-xnli-multilingual-nli-2mil7 | 0.69 | 0.73 | 0.69 | 0.68 |
| | Gemma3 | **0.88** | **0.88** | **0.88** | **0.88** |
| **E-CONAN-2** | MarBERT-XNLI-Tuned | 0.76 | 0.75 | 0.74 | 0.74 |
| | mDeBERTa-v3-base-xnli-multilingual-nli-2mil7 | 0.84 | 0.83 | 0.79 | 0.81 |
| | Gemma3 | **0.95** | **0.96** | **0.94** | **0.95** |

### d. Zero-Shot Cross-Dataset Transfer Analysis

To demonstrate the utility, robustness, and challenging nature of our proposed E-CONAN-3 and E-CONAN-2 benchmarks, we conduct a cross-dataset transfer experiment. We tested nine state-of-the-art pretrained models originally trained on standard datasets (MNLI and XNLI). These models are evaluated directly on our newly introduced benchmarks, XNLI, and ArNLI as reference baselines. The empirical results are detailed in Table 6 and Table 7, respectively.

The evaluation yields several key insights regarding how well existing SOTA models transfer to data distributions:

- The Generalization Gap: While multilingual models like MoritzLaurer/mDeBERTa-v3-base-mnli-xnli perform exceptionally well on standard XNLI (achieving an accuracy of 0.87), their performance drops significantly to 0.70 when evaluated on E-CONAN-3. This performance degradation clearly indicates that our dataset introduce unique contextual distributions and complex semantic challenges that standard XNLI training does not fully capture.
- Architecture Limitations: Conversely, English-centric models (such as RoBERTa and BART variants) exhibit very poor cross-dataset efficacy, failing to cross the 0.36 accuracy threshold on E-CONAN-3. Meanwhile, robust multilingual pipelines (mDeBERTa-v3 and ernie-m-base) retain moderate generalization capabilities (scoring 0.71 and 0.68, respectively), yet still reveal substantial room for improvement.

These cross-dataset transfer results validate the distinct value of E-CONAN-2 and E-CONAN-3 as benchmarks. By testing models without any supervised fine-tuning or parameter tuning, our evaluation successfully exposes the true boundaries of current state-of-the-art models, proving that high performance on traditional NLI datasets does not guarantee seamless adaptation to new, specialized domains.

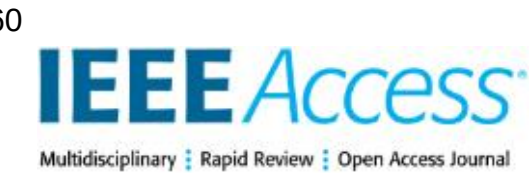

## VI. Error Analysis

We conducted a detailed error analysis to better understand models' performance. This included a qualitative evaluation via manual inspection to identify error patterns, as well as a quantitative analysis to measure the frequency of those patterns.

### a. Qualitative Error Analysis

To investigate the root causes of models' failure, we conducted a manual qualitative analysis, identifying four primary failure modes that persist across both pretrained models and LLMs. (See representative samples in Table 11)

- **Lexical Overlap and Heuristic Bias**: most models show a strong lexical overlap heuristic bias, frequently predicting Entailment when the premise text (T) and hypothesis (H) share high-frequency tokens, regardless of their logical relationship. For instance, when T describes "two dogs running" (كلبان يركضان) and H adds a specific location like "in the garden" (في الحديقة), the model predicts Entailment. This suggests that the model relies on bag-of-words similarity rather than recognizing that H contains new, unverified information, which logically requires a Neutral label.
- **Entity Mismatch and Lack of World Knowledge**: A significant failure mode where models fail to distinguish between conflicting entities despite sharing the same action verbs. This is frequently observed across numerous instances; for example, where the model predicts Entailment for T: "Assad's children spent their vacation in a Russian camp" and H: "Assad's children spent their vacation in America". The model focus on the event type ("vacation") while ignoring the geographic contradiction between "Russia" and "America". This confirms a lack of world knowledge, where named entities are treated as identical tokens rather than distinct locations.
- **Linguistic Challenges:** The complex morphology of Arabic introduces specific challenges regarding agreement and semantic:
  - **Gender Disagreement (Morphological Mismatch):** a frequent pattern observed in many cases where models often fail to recognize that the gender shift implies a different set of entities. For instance, when T involves "two young men sitting" (شابان يجلسان) and H involves "two young women sitting" (شابتان تجلسان), most models wrongly predict Entailment instead of Neutral.
  - **Quantifier and Numerical Constraints**: most models struggle with the logical boundaries of quantifiers. For example, the transition from a general plural in T ("children in the park") to a specific count in H ("three children in the park") is frequently misclassified as Entailment, ignoring the lack of numerical evidence in the premise text.
  - **Antonymy and Negation Blindness**: many models demonstrate antonym blindness in the presence of high lexical overlap. For example, T: "lifting the curfew" (رفع حظر التجول) and H: "continuation of the curfew" (استمرار حظر التجول) is often misidentified as Entailment due to the shared "curfew" token, despite the explicit semantic contradiction.
- **Topical Hallucination**: We found that models often get distracted by shared topics and emotions, leading to semantic over-generalization or over-specification. If two sentences share a keyword and an emotion, the model tends to mark them as Entailment, even if the facts don't align. For instance, some models link a premise text about a reporter's work to a hypothesis about a reporter's death just because they feel related, failing to see that there is no logical connection between the two. This tendency leads the model to jump to conclusions by adding specific details that don't exist in the text. For example, many models predict Entailment for T: "two people riding bikes" and H: "two people are in competitive race" as the model associates biking with racing, it assumes the specific context is true, failing to realize that it has moved from a general fact to an unverified guess.

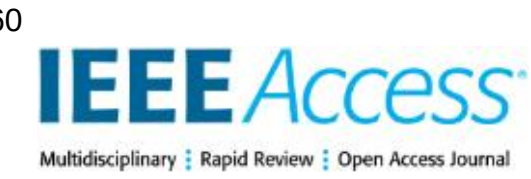

TABLE 11
QUALITATIVE ERROR ANALYSIS

| Prediction | Class | Hypothesis (H) | Premise Text (T) |
|---|---|---|---|
| Entailment | Neutral | وفاة صحفي بعد صراع مع المرض | وفاة صحفي كتب عن المرتزقة الروس في سوريا في ظروف غامضة |
| Entailment | Contradiction | استمرار حظ التجول فى البصرة | رفع حظ التجول في البصرة |
| Entailment | Contradiction | أبناء الأسد قضوا العطلة في أمريكا | معسكر روسي يستضيف أبناء الأسد في عطلتهم |
| Entailment | Contradiction | إسرائيل: هناك فرق كبير بين الدولة اللبنانية وحزب الله | إسرائيل: لن نفرق بين الدولة اللبنانية وحزب الله في أي حرب في المستقبل |
| Entailment | Neutral | شابان يجلسان | شابتان تجلسان |
| Contradiction | Neutral | شابان يمشيان | شخصان يمشيان |
| Entailment | Neutral | طفلان يركضان | شخصان يركضان |
| Entailment | Neutral | كلبان يركضان في الحديقة | كلبان يركضان |
| Entailment | Neutral | مجموعة من الناس يتسوقون في سوق بالشارع يبيع الفاكهة | مجموعة أشخاص يمشون في الخارج |
| Entailment | Neutral | شخصان في سباق | شخصان يركبان الدراجة |
| Entailment | Neutral | طفلان يلعبان بالكرة | شخصان يمارسان الرياضة |
| Entailment | Neutral | ثلاثة أطفال في الحديقة | أطفال في الحديقة |
| Neutral | Contradiction | الألغاز ممتعة وتثقيفية أيضاً. | الألغاز مملة ولا تُعلمك شيئًا. |

### b. Quantitative Error Analysis

To better understand the characteristics of the challenges in Arabic NLI, we conducted a statistical error analysis focusing on representative models that showed significant performance gaps. Specifically, we analyzed MoritzLaurer/MiniLM-L6-mnli (representing cross-lingual pretrained models) and Allam (representing Large Language Models). By examining a representative sub-sample of errors from these two models, we identified frequent failure modes that highlight the divergence between traditional fine-tuned models and zero-shot LLM performance (see Figure 14).

Results show that while pretrained models are often misled by lexical overlap, LLMs are often misled by topical familiarity. This suggests that for Arabic NLI, both architectures prioritize similarity over logical consistency. Furthermore, both approaches suffer from a bias toward the Entailment class.

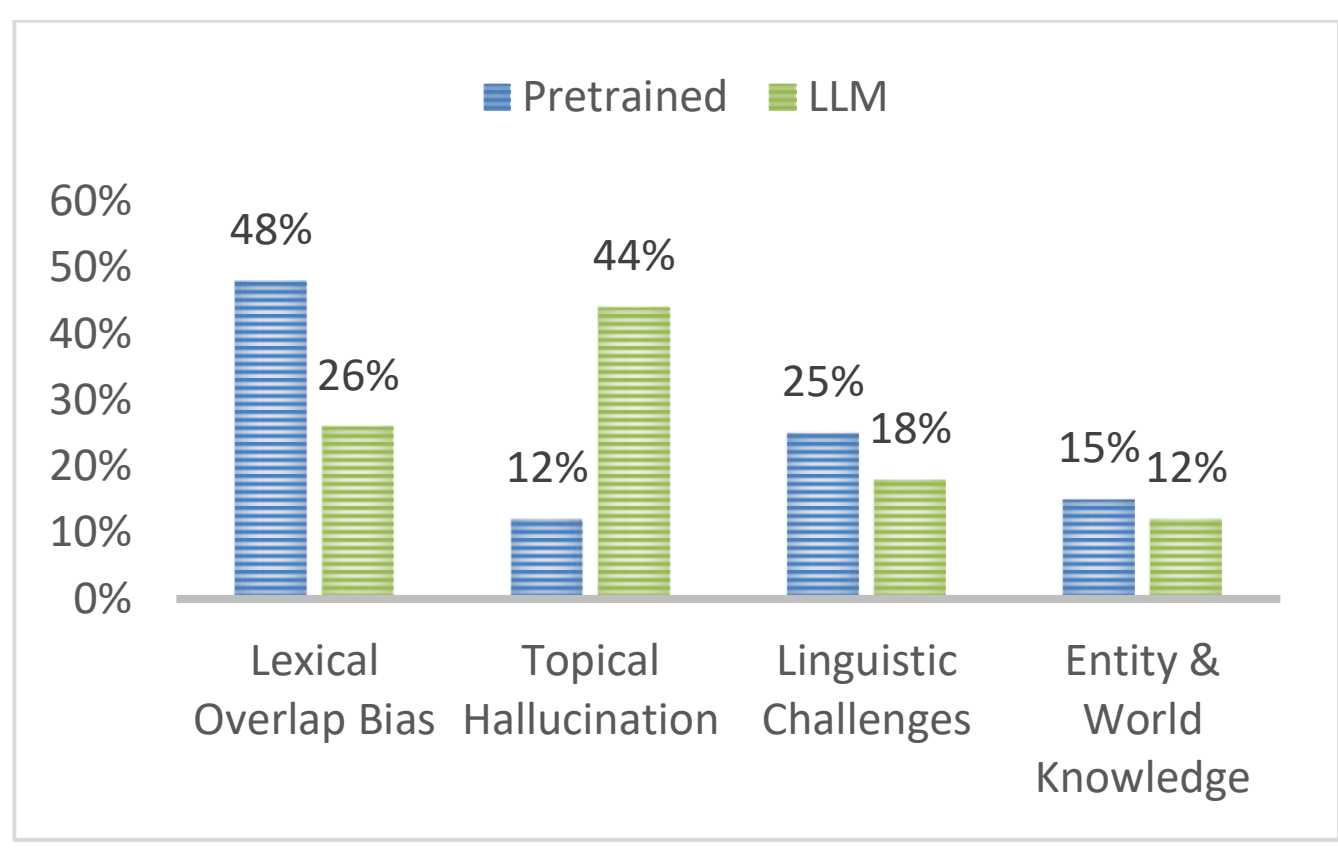

*Figure14: Quantitative Error Analysis*

### c. Class-Wise Performance and Error Patterns

While overall accuracy provides a useful snapshot of models' performance, it does not tell how well the models handle specific types of linguistic relationships. To gain a deeper understanding, we studied class-wise performance using the classification reports (Table 12) and confusion matrices (Figure 15). This analysis highlights models' strengths, common error patterns, and the robustness of proposed benchmarks across different linguistic relations.

- **Challenge of Neutral and Contradiction Classes:**

Despite the overall strength of encoder-based models, a closer inspection of class-wise results highlights that the neutral and contradiction classes remain particularly challenging across all architectures. As shown in the confusion matrices (Figure 15), there is consistent misclassification between these two categories, with a noticeable tendency for both contradiction and entailment instances to be predicted as neutral.
Although Gemma3 achieves competitive overall accuracy, a closer examination of class-wise performance reveals a tendency to favor the majority class in more challenging inference scenarios. This behavior is particularly evident on ArSNLI and ArNLI, where the model frequently predicts the Neutral label. Consequently, while Neutral recall remains relatively high, performance on Contradiction and Entailment decreases, indicating difficulty in distinguishing between semantic classes in harder cases.

This difficulty can be attributed to several factors:

- Subjective annotation boundaries: the datasets integrated into E-CONAN-3 rely heavily on crowdsourced annotations, where distinguishing between contradiction and neutral cases is often subjective. In particular, determining whether missing information implies non-contradiction

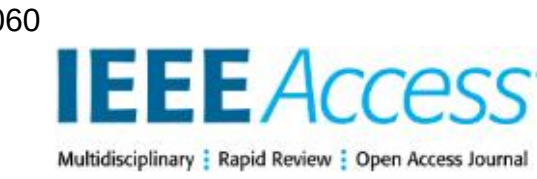

(neutral) or direct conflict (contradiction) introduces ambiguity that is difficult for models to resolve consistently.

- Linguistic and translation artifacts: since benchmarks such as XNLI and SICK include translated or machine-translated Arabic data, some of the original pragmatic and cultural cues may be lost. This can reduce the clarity of semantic contradictions, leading models to favor the neutral class when evidence for a strong conflict is not explicit.
- Lexical overlap bias**:** models also tend to rely on surface-level lexical overlap as a heuristic. When premise and hypothesis share similar vocabulary but differ due to negation or indirect contextual shifts, models often struggle to capture the semantic difference, resulting in frequent confusion between contradiction and neutral labels.

- **Robustness, Semantic Diversity, and Fine-Tuning Potential of E-CONAN-3**

The consistent zero-shot performance observed across all evaluated models on E-CONAN-3 suggests that the benchmark offers advantages over individual existing datasets. Whereas several prior benchmarks exhibit label-specific biases that can encourage shortcut-based predictions, the aggregation of multiple data sources within E-CONAN-3 appears to mitigate these effects. By combining diverse datasets without extensive filtering, E-CONAN-3 increases contextual and linguistic diversity while reducing the influence of dataset-specific stylistic patterns. This broader coverage is reflected in more balanced class-wise performance and clearer diagonal structures in the confusion matrices.

Despite these improvements, E-CONAN-3 still inherits certain limitations from its underlying source datasets. In particular, some degree of domain imbalance and topic distribution bias remains, as the benchmark is skewed toward specific structural genres. As shown in our domain breakdown (see Figure 4), the data is mainly composed of news (35%) and simple everyday (30%) contexts, while social media (20%) and multi-genre (15%) settings are comparatively underrepresented. In addition, since E-CONAN-3 is built from aggregated datasets such as SNLI, XNLI, and SICK, residual effects of translation bias and crowdsourcing-related annotation bias may still be present. Although combining multiple datasets can help reduce the impact of individual dataset biases compared to single-source benchmarks, these inherited limitations should still be taken into account when interpreting model generalization performance.

Acknowledging these factors, E-CONAN-3 still provides a strong foundation for future fine-tuning efforts, potentially promoting better generalization and more robust semantic understanding in Arabic Natural Language Inference models.

## VII. Conclusion

This paper addresses a critical gap in Arabic Textual Entailment and Natural Language Inference (NLI) benchmarks by introducing E-CONAN datasets composed of diverse sentence pairs from multiple sources. E-CONAN contains both 2-way (RTE) named E-CONAN-2 and 3-way (NLI) named E-CONAN-3. E-CONAN sources are combination of automatically translated pairs, human-validated machine-translated pairs, hand-crafted pairs, and news-based pairs, offering wide-ranging evaluation datasets from different types of texts. To evaluate E-CONAN, we conducted zero-shot classification on nine state-of-the-art multilingual pre-trained models. Results on E-CONAN datasets were compared with results on ArNLI and XNLI datasets. The challenging nature of the ArNLI dataset, characterized by its high-quality human-validated and manually crafted data, led to the lowest observed performance. On the other hand, the XNLI dataset, likely benefiting from its commonness in model training, yielded the highest results. E-CONAN demonstrated an intermediate difficulty level making it a valuable tool for assessing model generalization. Its diverse composition, unlike the singular nature of ArNLI and the potentially biased machine-translated origins of XNLI, provides a more balanced and representative evaluation. As for best pretrained models in this paper experiment, results show that mDeBERTa model consistently outperformed all other models on all datasets, achieving accuracies of 0.71 and 0.86 on E-CONAN-3 and XNLI datasets, respectively. This highlights the effectiveness of large-scale multilingual pre-training and targeted fine-tuning dataset.

Moreover, we evaluated 5 LLMs on E-CONAN-3 dataset. Best results were achieved by Gemma with an accuracy of 0.68. Additionally, after deep investigation we found that most errors were between neutral and contradiction classes. Thus, we calculated results on E-CONAN-3 as 2-way, where best results were achieved by Gemma and Qwen with an accuracy of 0.95, 0.94 respectively.

In addition, we incorporated MARBERT as a representative Arabic-specific baseline and conducted performance evaluation comparison to demonstrate how Arabic-specific models scale against cross-lingual and

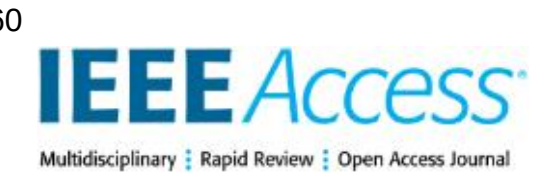

LLM-based approaches on the E-CONAN benchmarks. Moreover, we conducted detailed qualitative and quantitative error analysis to analyze frequent error patterns. Results show that while pretrained models are often misled by lexical overlap, LLMs are often misled by topical familiarity.

In summary, E-CONAN datasets could be a good contribution to the Arabic NLI research domain, offering robust and diverse evaluation datasets. The results highlight the importance of diverse, high-quality benchmarks for accurately assessing and improving model generalization, particularly in resource-limited languages such as Arabic. We do hope that future researches make use of E-CONAN datasets to further refine multilingual models to enhance RTE and NLI performance in diverse linguistic contexts.

While current E-CONAN benchmarks provide a robust evaluation for Modern Standard Arabic (MSA), regional Arabic dialects are not currently represented. Future work will involve datasets expansion to include various dialects, ensuring broader coverage of the linguistic variations characteristic of Arabic language.

## Abbreviations

NLI: Natural Language Inference
RTE: Recognizing Textual Entailment

## Data Availability

E-CONAN datasets are available on Hugging Face[17].

[17] https://huggingface.co/datasets/KhloudJ/E-CONAN

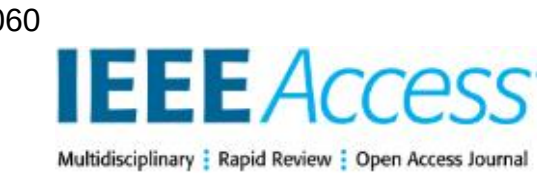

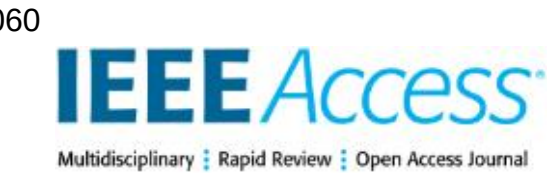

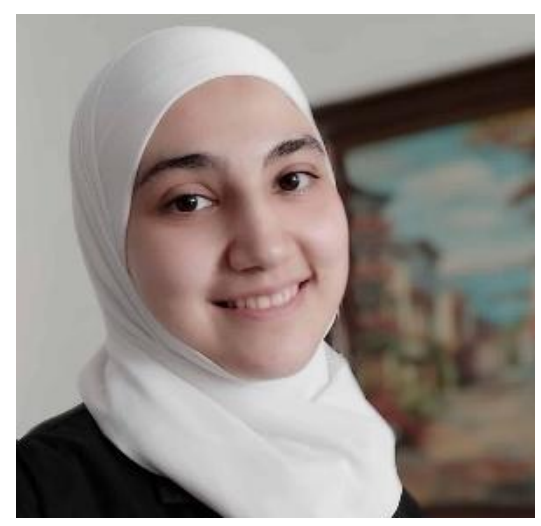

**Khloud Al Jallad** received the B.S. degree in informatics engineering from Arab International University, Syria, in 2014, and the M.D. degree in Big Data from Higher Institute for Applied Sciences and Technology (HIAST), Syria, in 2019. Currently a PhD candidate at HIAST, Syria. From 2017 till now, she has been working as a researcher and lecturer. Her research interests include NLP, DL, ML.

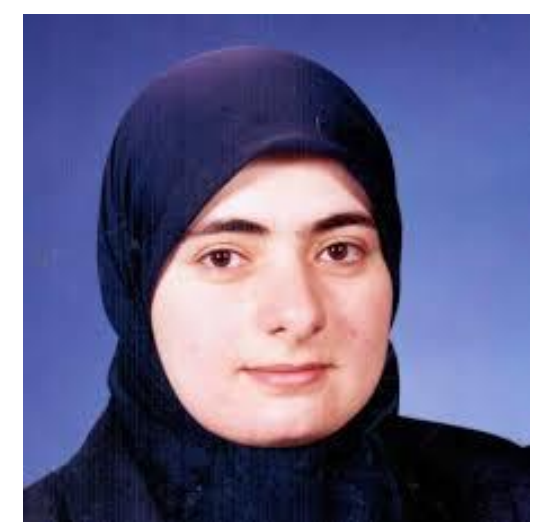

**Ghaida Rebdawi.** received the DEA and Ph.D. degrees in applied automatic and informatics from the Institut National des Sciences Appliquées de Lyon, France, in 1987 and 1990, respectively. In 1991, she joined the Informatics Department, at the Higher Institute for Applied Sciences and Technology, as a Research Assistant Professor. She is actually a Full Professor and a Research Director, at the forementioned institute. She coauthored many academic books in computer science, and more than 20 articles. Her research interests include software engineering, requirements engineering, and natural language processing.

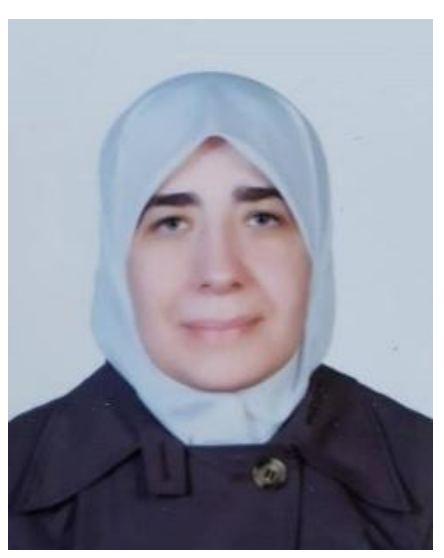

**Nada Ghneim** received the Information Technology Engineering degree from Higher Institute for Applied Sciences and Technology (HIAST), Damascus, Syria, 1991, and the Postgraduate Degree (DEA) in Artificial Intelligence (Image, Robotics, Vision), from the National High School of Computer Science and Applied Mathematics in Grenoble (ENSIMAG), France, 1993, and a PhD degree in Language Sciences (Speech Communication) from “Institut de la Communication Parlée”-Stendhal University, Grenoble, France, 1997. Associate Professor at the Faculty of Informatics \& Communication Engineering, at the Arab International University (AIU), Daraa, Syria. She is also a Lecturer, at HIAST, the Syrian Virtual University and the Information Technology Engineering Faculty at Damascus University. Professor Ghneim’s research interests include Artificial Intelligence and Natural Language Processing. She has many publications in Conferences, workshops, journals and books, mainly focusing on Arabic language processing at different modalities (speech, text and image), such as Text-to-Speech, Image Captioning, Sentiment Analysis, Inference detection, …

Shared Data Sources

ArNLI

ArSNLI

ArXNLI

AnsStance

ArEntail

Data Collection & Merging

Label Normalization

AnsStance → NLI Mapping

agree → entailment

disagree → contradiction

others → neutral

3-Way → 2-Way (E-CONAN-2)

entail → entail

contradiction → not-entail

neutral → not-entail

Label Mapping Verification

Final E-CONAN-3

NLI (3-way)

Entailment / Neutral / Contradiction

Final E-CONAN-2

RTE (2-way)

Entail / Not-Entail

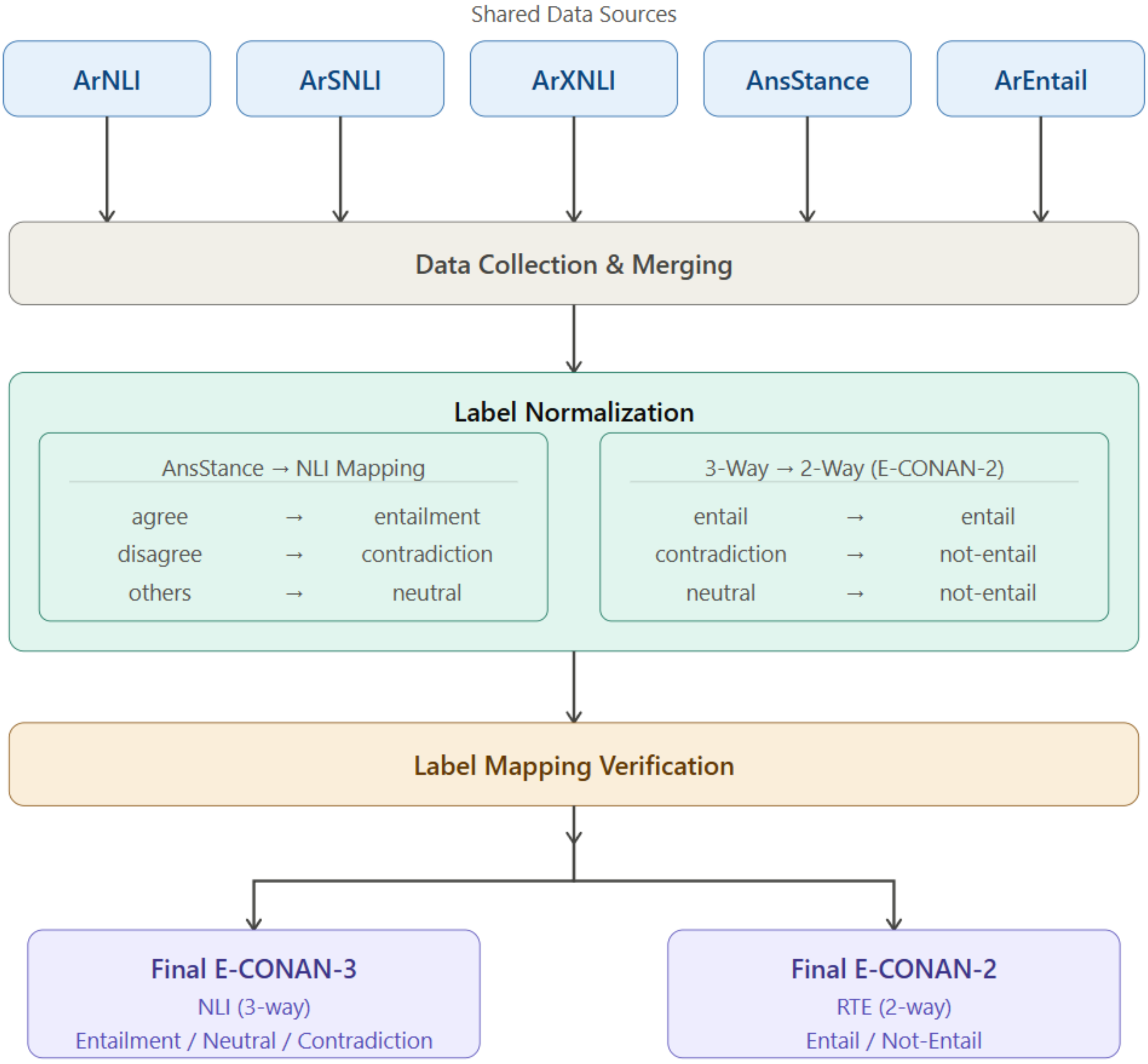

**Figure2: E-CONAN Datasets Construction Pipeline**

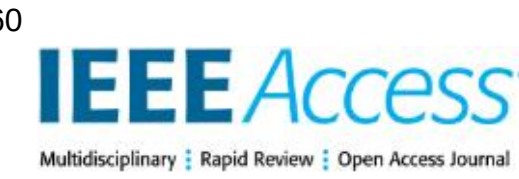

mDeBERTa -V3-Base-XNLI-Multilingual-NLI-2mil7
Facebook Bart Large MNLI
Ernie-M-Base-MNLI-XNLI
Deberta-V3-Base-MNL
mDeBERTa-V3-Base-MNLI-XNLI
Multilingual-MiniLMv2-L12-MNLI-XNLI
MiniLM-L6-MNLI
FacebookAI RoBERTa-Large-MNLI
Multilingual-MiniLMv2-L6-MNLI-XNLI

Text
Hypothesis
Logical Relation

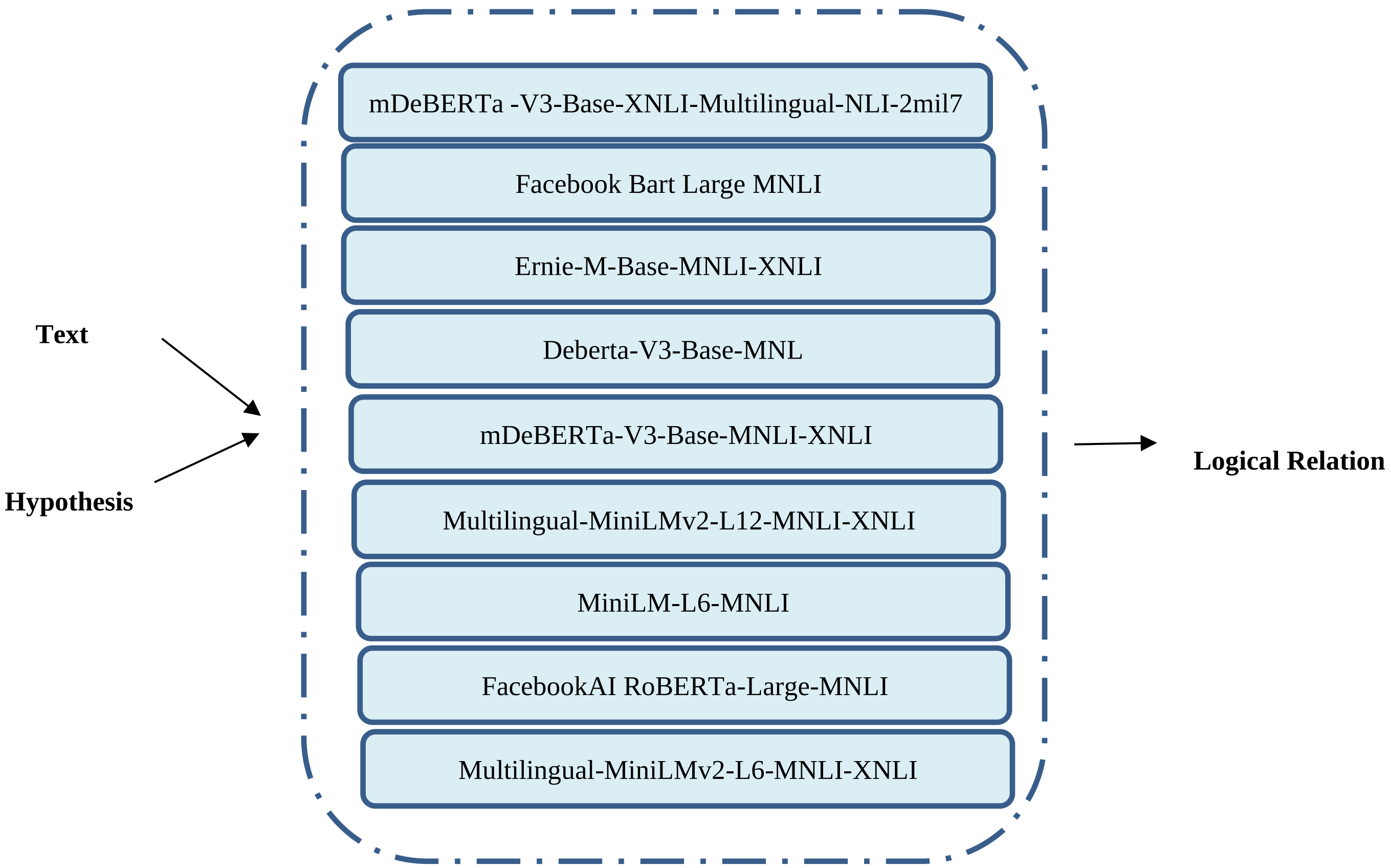

**Figure6**: **Proposed Evaluation of Cross-lingual Baseline Models**

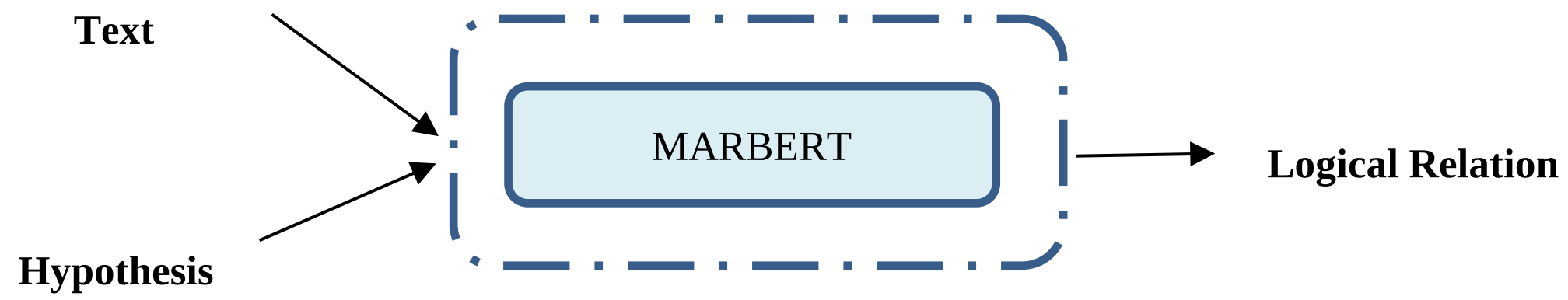

**Figure7**: **Proposed Evaluation of Arabic Baseline Model**

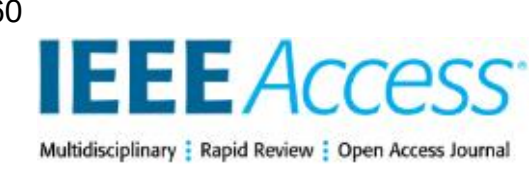

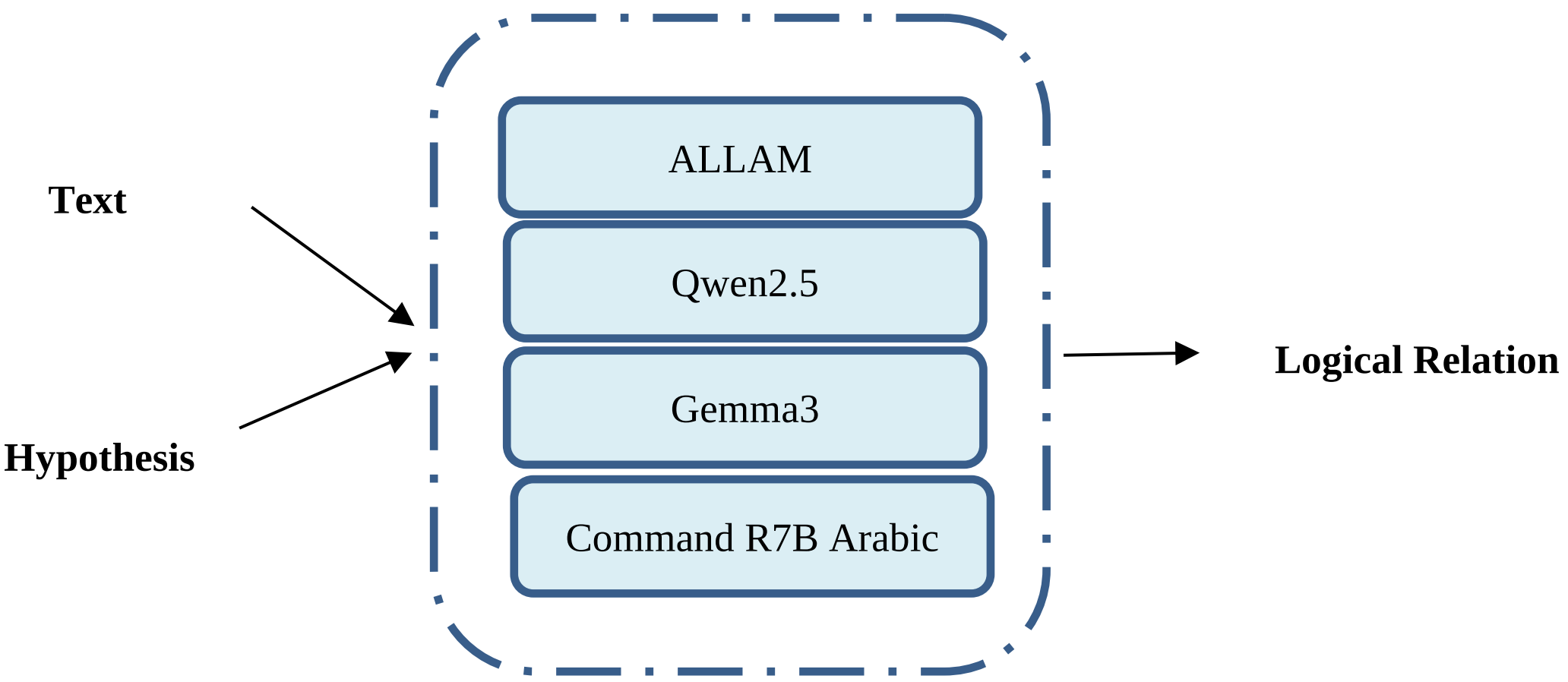

**Figure8: Proposed Evaluation of LLMs**

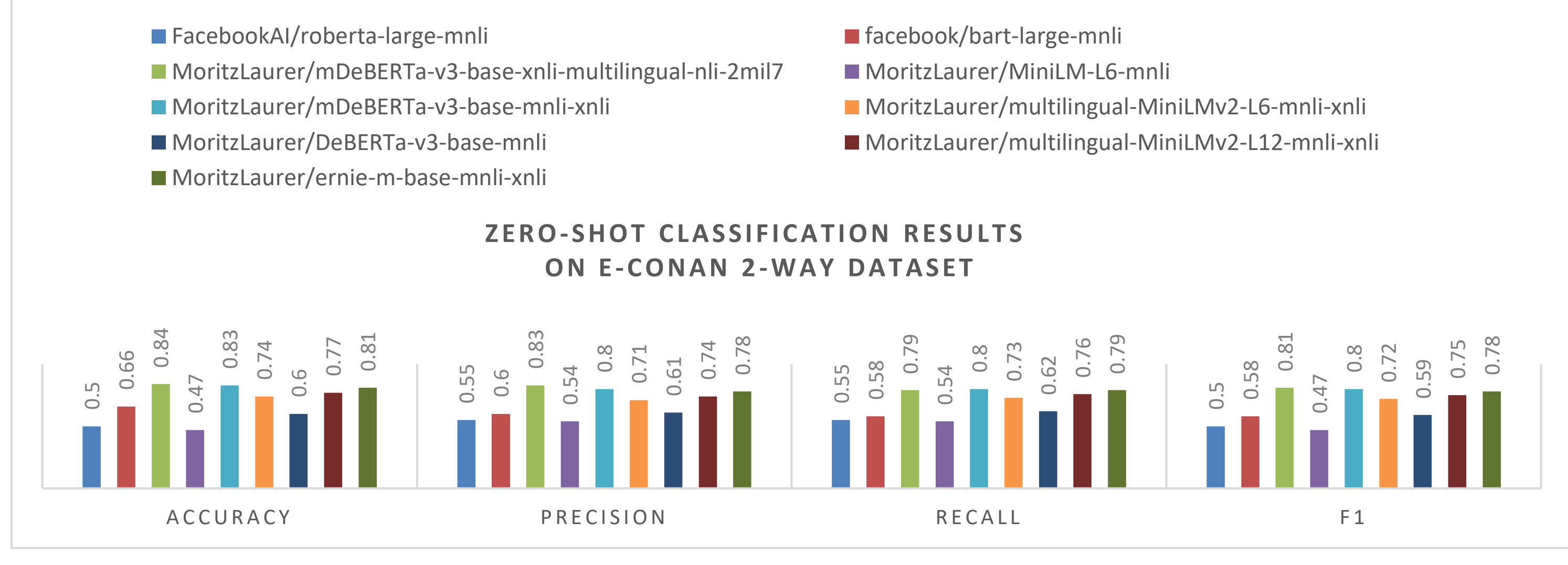

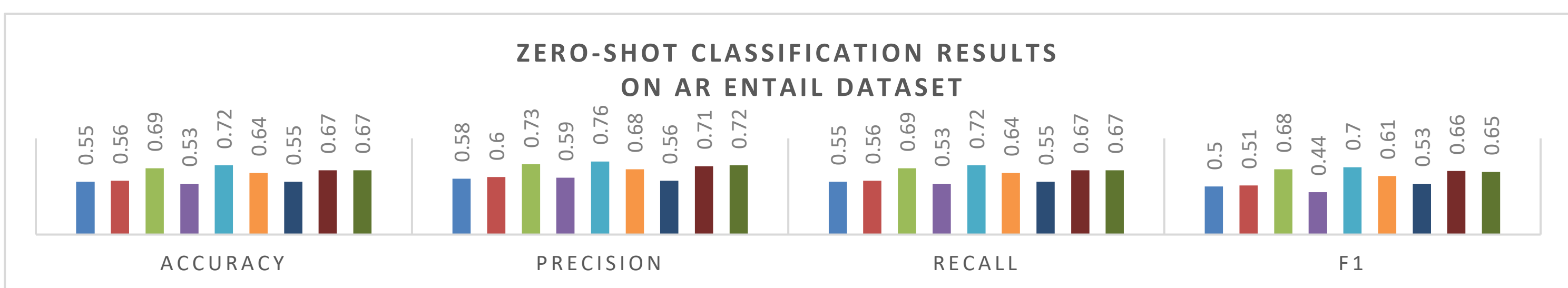

**Figure9: Zero-Shot Classification Results on ArEntail, E-CONAN-2 Datasets**

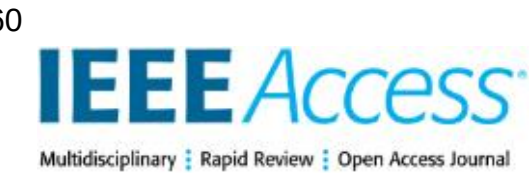

FacebookAI/roberta-large-mnli
facebook/bart-large-mnli
MoritzLaurer/mDeBERTa-v3-base-xnli-multilingual-nli-2mil7
MoritzLaurer/MiniLM-L6-mnli
MoritzLaurer/mDeBERTa-v3-base-mnli-xnli
MoritzLaurer/multilingual-MiniLMv2-L6-mnli-xnli
MoritzLaurer/DeBERTa-v3-base-mnli
MoritzLaurer/multilingual-MiniLMv2-L12-mnli-xnli
MoritzLaurer/ernie-m-base-mnli-xnli

ZERO-SHOT CLASSIFICATION RESULTS ON E-CONAN-3 DATASETS

ACCURACY: 0.34 0.36 0.71 0.33 0.7 0.61 0.46 0.64 0.68
PRECISION: 0.3 0.36 0.72 0.38 0.7 0.61 0.47 0.64 0.67
RECALL: 0.34 0.36 0.71 0.34 0.7 0.61 0.46 0.64 0.68
F1: 0.27 0.32 0.71 0.25 0.7 0.61 0.44 0.64 0.67

ZERO-SHOT CLASSIFICATION RESULTS ON XNLI DATASET

ACCURACY: 0.36 0.38 0.86 0.35 0.87 0.77 0.57 0.81 0.85
PRECISION: 0.42 0.45 0.87 0.4 0.87 0.77 0.57 0.81 0.85
RECALL: 0.36 0.38 0.86 0.35 0.87 0.77 0.57 0.81 0.85
F1: 0.3 0.31 0.86 0.27 0.87 0.77 0.57 0.81 0.85

ZERO-SHOT CLASSIFICATION RESULTS ON ARNLI DATASET

ACCURACY: 0.23 0.25 0.55 0.26 0.49 0.46 0.37 0.44 0.48
PRECISION: 0.17 0.34 0.55 0.15 0.5 0.47 0.39 0.46 0.50
RECALL: 0.28 0.31 0.59 0.32 0.57 0.53 0.47 0.54 0.58
F1: 0.19 0.27 0.54 0.2 0.49 0.46 0.36 0.45 0.49

ZERO-SHOT CLASSIFICATION RESULTS ON TEST XNLI DATASET

ACCURACY: 0.37 0.38 0.79 0.35 0.80 0.69 0.57 0.73 0.78
PRECISION: 0.43 0.45 0.80 0.45 0.80 0.69 0.57 0.73 0.78
RECALL: 0.37 0.38 0.79 0.35 0.80 0.69 0.57 0.73 0.78
F1: 0.31 0.31 0.79 0.27 0.80 0.69 0.57 0.73 0.78

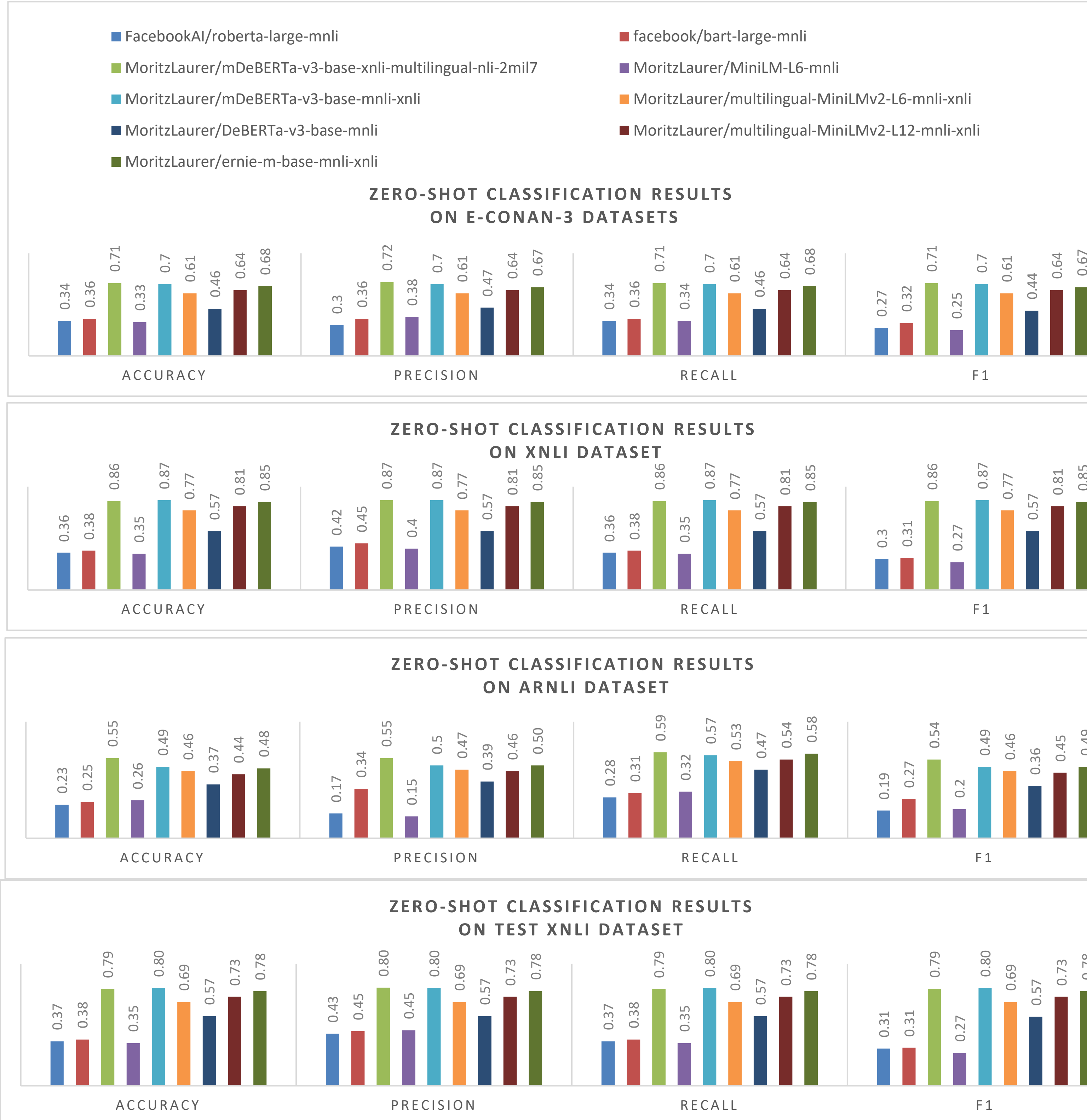

**Figure10: Results on E-CONAN-3, ArNLI and XNLI Datasets**

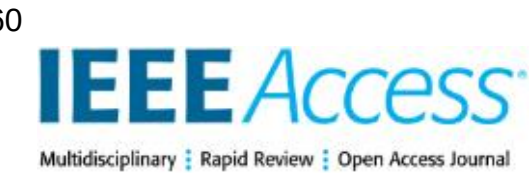

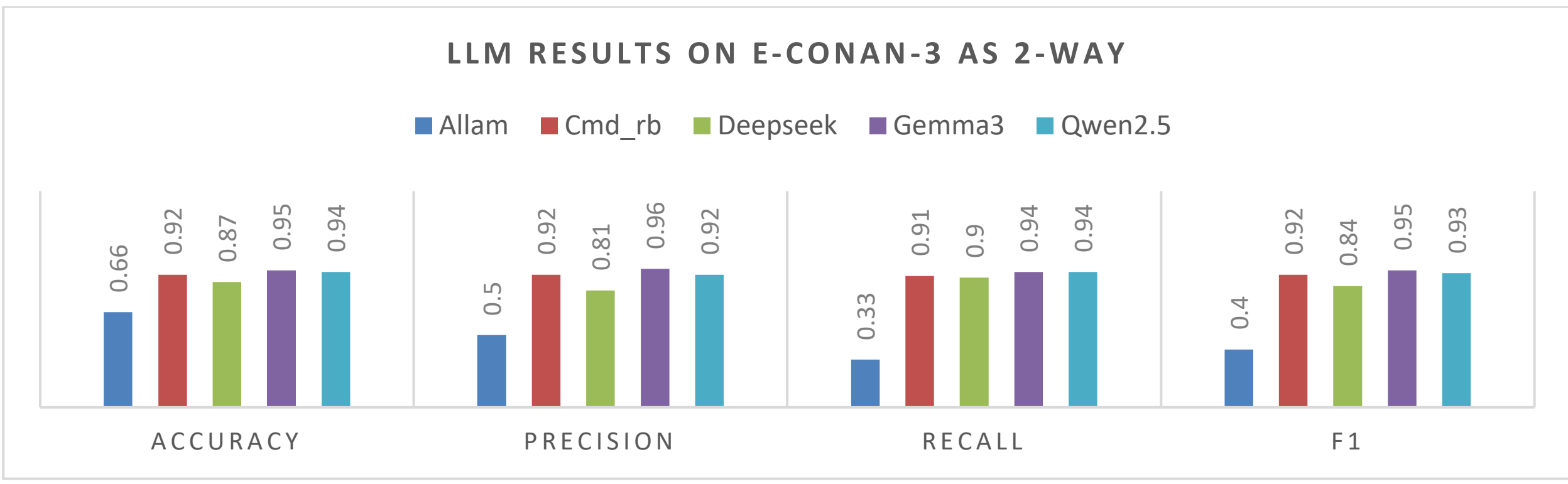

Figure11: **LLM Zero-Shot Classification on E-CONAN-3 as 2-way**

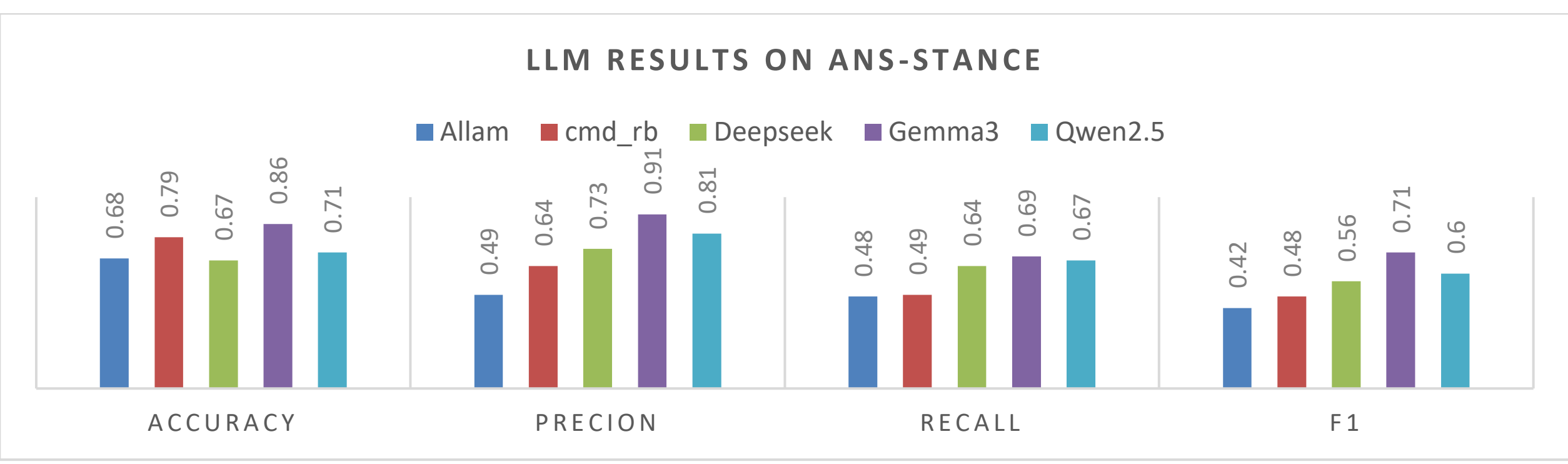

Figure12: **LLM Zero-Shot Classification on Ans-Stance Dataset**

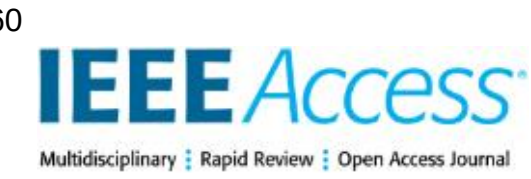

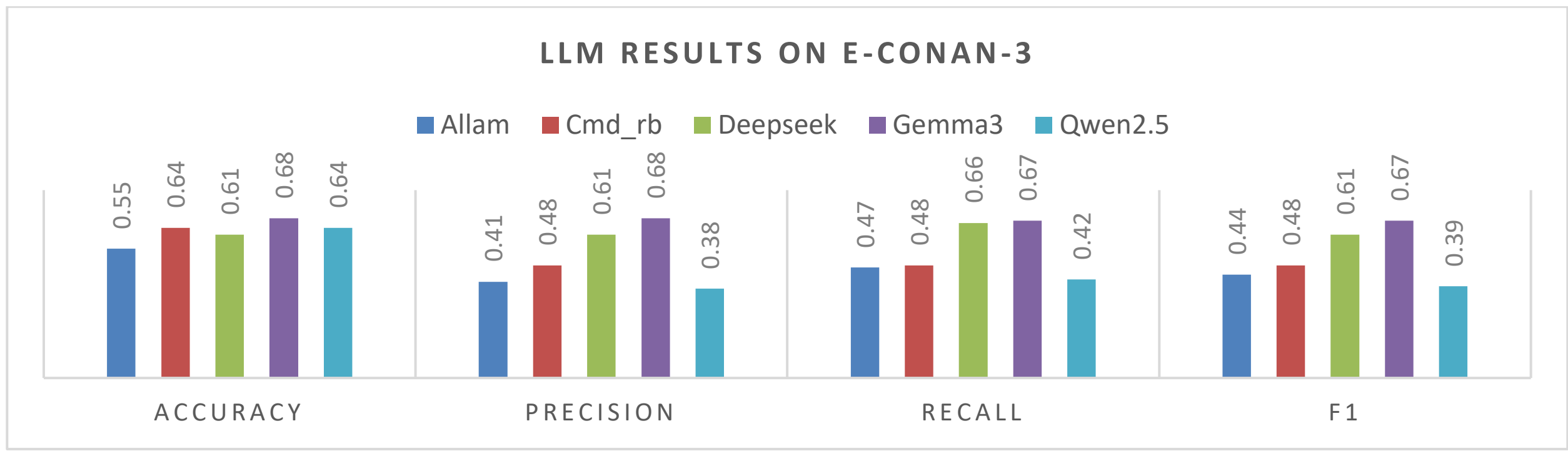

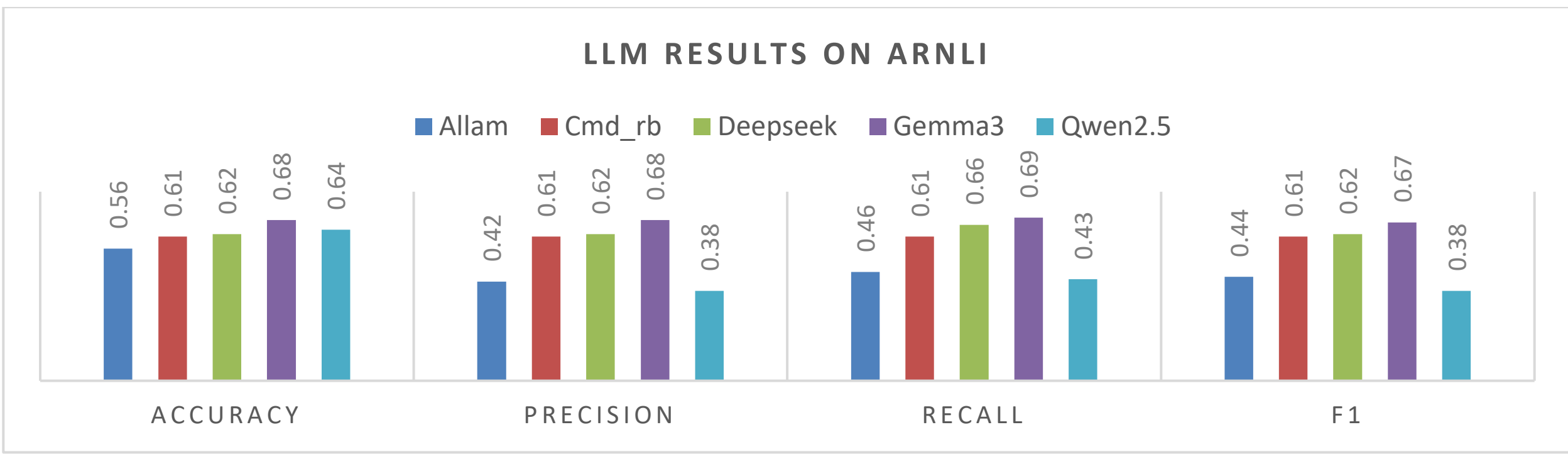

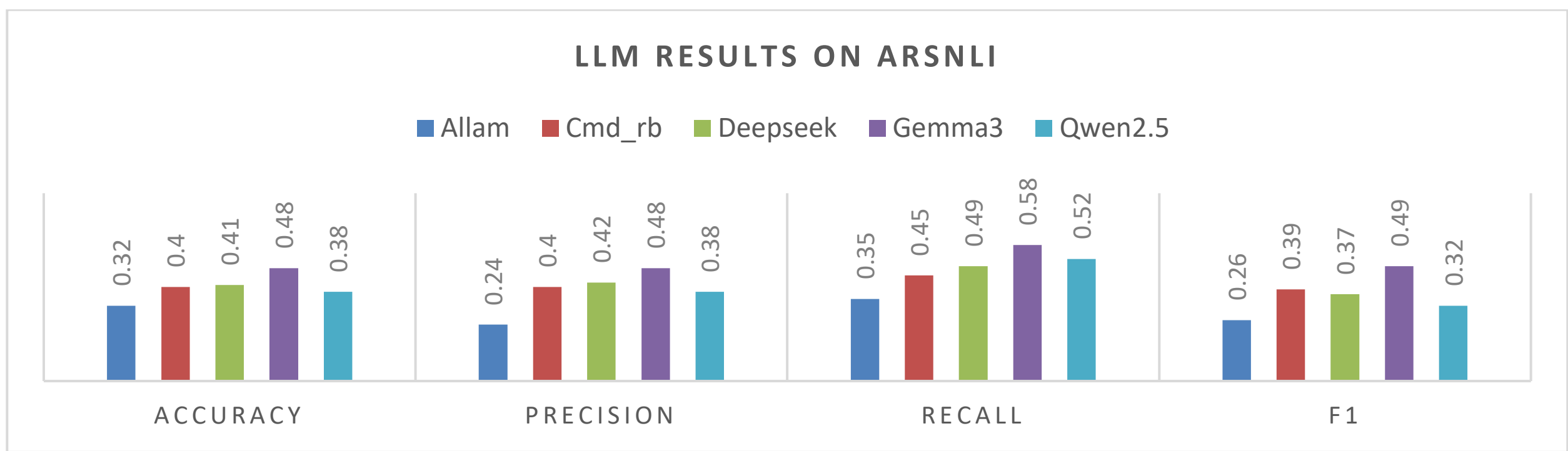

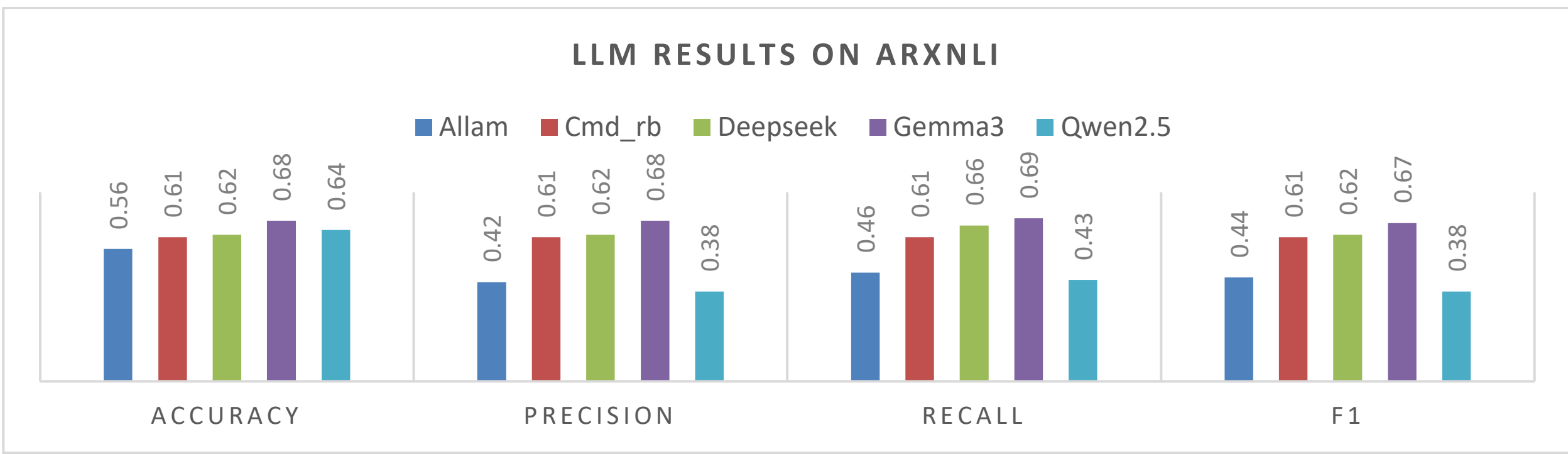

**Figure13: LLM Zero-Shot Classification on E-CONAN-3 and SoTA Benchmarks**

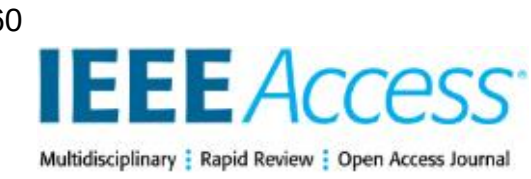

TABLE 12
CLASSIFICATION REPORT OF E-CONAN AND SOTA BENCHMARKS.

| Dataset Name | Model | Precision | | | Recall | | | F1 | | |
|---|---|---|---|---|---|---|---|---|---|---|
| | | Contradiction | Entailment | Neutral | Contradiction | Entailment | Neutral | Contradiction | Entailment | Neutral |
| **ArNLI** | **MarBERT-XNLI-Tuned** | 0.28 | 0.56 | 0.63 | 0.78 | **0.66** | 0.20 | 0.41 | **0.61** | 0.31 |
| | **mDeBERTa-v3-base-xnli-multilingual-nli-2mil7** | 0.38 | 0.63 | 0.63 | **0.75** | 0.52 | 0.50 | 0.51 | 0.57 | 0.56 |
| | **Gemma3** | **0.55** | **0.77** | **0.71** | 0.70 | 0.42 | **0.84** | **0.62** | 0.54 | **0.77** |
| **ArXNLI** | **MarBERT-XNLI-Tuned** | 0.79 | 0.77 | 0.62 | 0.73 | 0.63 | 0.78 | 0.76 | 0.69 | 0.69 |
| | **mDeBERTa-v3-base-xnli-multilingual-nli-2mil7** | **0.91** | **0.91** | **0.78** | **0.88** | **0.81** | **0.89** | **0.90** | **0.86** | **0.83** |
| | **Gemma3** | 0.84 | 0.73 | 0.47 | 0.50 | 0.58 | 0.76 | 0.63 | 0.65 | 0.58 |
| **ArSNLI** | **MarBERT-XNLI-Tuned** | 0.51 | 0.43 | 0.33 | 0.50 | 0.32 | 0.42 | **0.50** | **0.37** | 0.37 |
| | **mDeBERTa-v3-base-xnli-multilingual-nli-2mil7** | **0.80** | 0.27 | **0.34** | **0.32** | 0.05 | 0.82 | 0.4 | 0.09 | **0.48** |
| | **Gemma3** | 0.32 | **0.45** | 0.32 | 0.05 | 0.05 | **0.89** | 0.09 | 0.09 | **0.48** |
| **AnsStance** | **MarBERT-XNLI-Tuned** | **0.96** | 0.75 | **0.14** | **0.76** | **0.85** | 0.71 | **0.85** | 0.80 | **0.24** |
| | **mDeBERTa-v3-base-xnli-multilingual-nli-2mil7** | **0.96** | 0.87 | 0.08 | **0.76** | 0.82 | 0.66 | **0.85** | 0.84 | 0.15 |
| | **Gemma3** | 0.90 | **0.95** | 0.07 | 0.59 | 0.78 | **1** | 0.71 | **0.86** | 0.14 |
| **E-CONAN-3** | **MarBERT-XNLI-Tuned** | 0.62 | 0.67 | 0.55 | 0.73 | 0.66 | 0.45 | 0.67 | 0.67 | 0.50 |
| | **mDeBERTa-v3-base-xnli-multilingual-nli-2mil7** | 0.75 | **0.80** | **0.6** | **0.78** | **0.67** | 0.68 | **0.76** | **0.73** | **0.63** |
| | **Gemma3** | **0.76** | 0.79 | 0.50 | 0.54 | 0.53 | **0.81** | 0.63 | 0.64 | 0.62 |

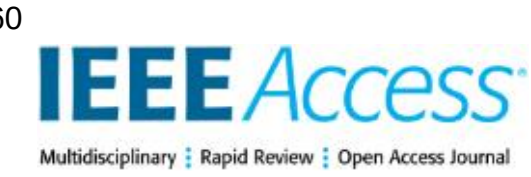

**ArNLI**

|  | Actual Label | Predicted: Contradiction | Predicted: Entailment | Predicted: Neutral |
|---|---|---|---|---|
| Gemma3 | Contradiction | 748 | 103 | 217 |
| Gemma3 | Entailment | 210 | 813 | 905 |
| Gemma3 | Neutral | 406 | 144 | 2801 |
| MarBERT | Contradiction | 829 | 143 | 96 |
| MarBERT | Entailment | 354 | 1278 | 296 |
| MarBERT | Neutral | 1800 | 872 | 679 |
| mDeBERTa | Contradiction | 805 | 90 | 173 |
| mDeBERTa | Entailment | 126 | 997 | 805 |
| mDeBERTa | Neutral | 1189 | 490 | 1672 |

**ArSNLI**

|  | Actual Label | Predicted: Contradiction | Predicted: Entailment | Predicted: Neutral |
|---|---|---|---|---|
| Gemma3 | Contradiction | 23 | 1 | 405 |
| Gemma3 | Entailment | 28 | 23 | 399 |
| Gemma3 | Neutral | 21 | 27 | 386 |
| MarBERT | Contradiction | 215 | 68 | 146 |
| MarBERT | Entailment | 81 | 145 | 224 |
| MarBERT | Neutral | 128 | 124 | 182 |
| mDeBERTa | Contradiction | 138 | 14 | 277 |
| mDeBERTa | Entailment | 6 | 23 | 421 |
| mDeBERTa | Neutral | 28 | 48 | 358 |

**ArXNLI**

|  | Actual Label | Predicted: Contradiction | Predicted: Entailment | Predicted: Neutral |
|---|---|---|---|---|
| Gemma3 | Contradiction | 1248 | 48 | 1190 |
| Gemma3 | Entailment | 128 | 1431 | 926 |
| Gemma3 | Neutral | 116 | 471 | 1904 |
| MarBERT | Contradiction | 1819 | 154 | 513 |
| MarBERT | Entailment | 237 | 1561 | 687 |
| MarBERT | Neutral | 240 | 302 | 1949 |
| mDeBERTa | Contradiction | 2195 | 55 | 236 |
| mDeBERTa | Entailment | 90 | 2022 | 373 |
| mDeBERTa | Neutral | 127 | 154 | 2210 |

**AnsStance**

|  | Actual Label | Predicted: Contradiction | Predicted: Entailment | Predicted: Neutral |
|---|---|---|---|---|
| Gemma3 | Contradiction | 1391 | 54 | 929 |
| Gemma3 | Entailment | 154 | 1013 | 127 |
| Gemma3 | Neutral | 0 | 0 | 85 |
| MarBERT | Contradiction | 1797 | 360 | 217 |
| MarBERT | Entailment | 52 | 1095 | 147 |
| MarBERT | Neutral | 23 | 2 | 60 |
| mDeBERTa | Contradiction | 1793 | 162 | 419 |
| mDeBERTa | Entailment | 39 | 1055 | 200 |
| mDeBERTa | Neutral | 27 | 2 | 56 |

**E-CONAN-3**

|  | Actual Label | Predicted: Contradiction | Predicted: Entailment | Predicted: Neutral |
|---|---|---|---|---|
| Gemma3 | Contradiction | 3410 | 206 | 2741 |
| Gemma3 | Entailment | 520 | 3280 | 2357 |
| Gemma3 | Neutral | 543 | 642 | 5176 |
| MarBERT | Contradiction | 4660 | 725 | 972 |
| MarBERT | Entailment | 724 | 4079 | 1354 |
| MarBERT | Neutral | 2191 | 1300 | 2870 |
| mDeBERTa | Contradiction | 4931 | 321 | 1105 |
| mDeBERTa | Entailment | 261 | 4097 | 1799 |
| mDeBERTa | Neutral | 1371 | 694 | 4296 |

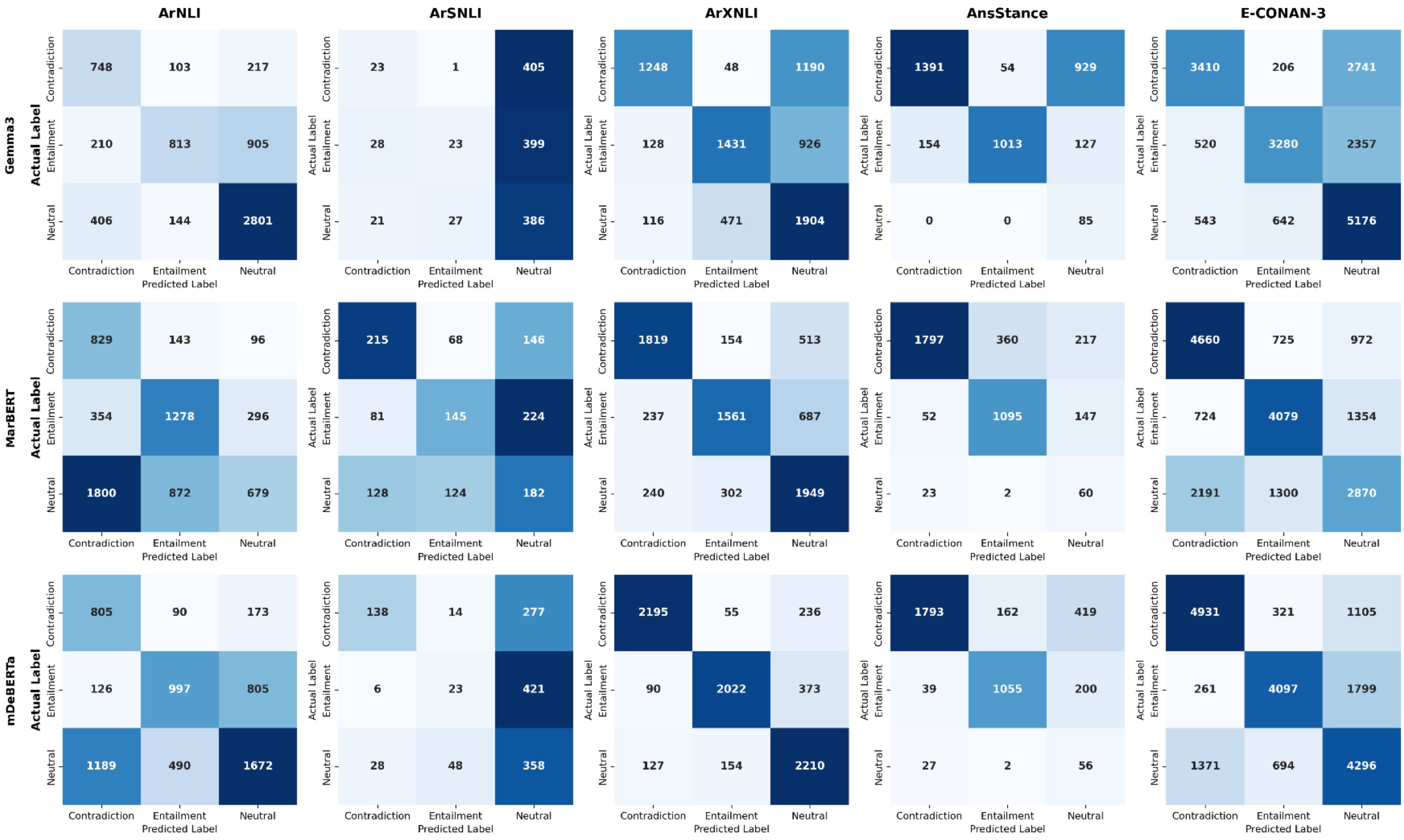

**Figure15: Confusion Matrix of E-CONAN and SoTA Benchmarks**